\documentclass[lettersize,journal]{IEEEtran}
\usepackage{amsmath,amsfonts,amssymb}
\usepackage{algorithm}
\usepackage[noend]{algpseudocode}
\usepackage{array}
\usepackage[caption=false,font=normalsize,labelfont=sf,textfont=sf]{subfig}
\usepackage{textcomp}
\usepackage{stfloats}
\usepackage{url}
\usepackage{verbatim}
\usepackage{graphicx}
\usepackage{cite}
\graphicspath{{./images/}}

\begin{document}

\title{A Dynamic Programming Framework for Optimal Planning of Redundant Robots Along Prescribed Paths With Kineto-Dynamic Constraints}

\author{Enrico~Ferrentino,~\IEEEmembership{Member,~IEEE,}
        Heitor~J.~Savino,~\IEEEmembership{Member,~IEEE,}
        Antonio~Franchi,~\IEEEmembership{Fellow,~IEEE,}
        and~Pasquale~Chiacchio,~\IEEEmembership{Senior~Member,~IEEE}
\thanks{E.\ Ferrentino and P.\ Chiacchio are with the Department of Computer and Electrical Engineering and Applied Mathematics (DIEM), University of Salerno, Fisciano, SA, 84084, Italy, e-mail: \{eferrentino,pchiacchio\}@unisa.it.}
\thanks{H.\ J.\ Savino is with Ambev Robotics Lab, Anheuser-Busch InBev, Jacareí, SP, 12334-480, Brazil, e-mail: heitor.savino@ambev.com.br.}
\thanks{A.\ Franchi is with the Robotics and Mechatronics Laboratory, Faculty of Electrical Engineering, Mathematics \& Computer Science, University of Twente, Enschede 7522NH, The Netherlands, and also with the
Department of Computer, Control and Management Engineering, Sapienza University of Rome, 00185 Rome, Italy; e-mails: a.franchi@utwente.nl and antonio.franchi@uniroma1.it}
}

\markboth{Accepted version, published  at https://doi.org/10.1109/TASE.2023.3330371 -- IEEE Transactions on Automation Science and Engineering}%
{Ferrentino \MakeLowercase{\textit{et al.}}: Time-Optimal Planning of Redundant Robots Along Prescribed Paths with Dynamic Programming}

\IEEEpubid{0000--0000/00\$00.00~\copyright~2023 IEEE}

\maketitle

\begin{abstract}
Offline optimal planning of trajectories for redundant robots along prescribed task space paths is usually broken down into two consecutive processes: first, the task space path is inverted to obtain a joint space path, then, the latter is parametrized with a time law. If the two processes are separated, they cannot optimize the same objective function, ultimately providing sub-optimal results. In this paper, a unified approach is presented where dynamic programming is the underlying optimization technique. Its flexibility allows accommodating arbitrary constraints and objective functions, thus providing a generic framework for optimal planning of real systems. To demonstrate its applicability to a real world scenario, the framework is instantiated for time-optimality on Franka Emika's Panda robot. The well-known issues associated with the execution of non-smooth trajectories on a real controller are partially addressed at planning level, through the enforcement of constraints, and partially through post-processing of the optimal solution. The experiments show that the proposed framework is able to effectively exploit kinematic redundancy to optimize the performance index defined at planning level and generate feasible trajectories that can be executed on real hardware with satisfactory results. 
\end{abstract}

\def\abstractname{Note to Practitioners}
\begin{abstract}
The common planning algorithms which consolidated over the years for generating trajectories for non-redundant robots are not adequate to fully exploit the more advanced capabilities offered by redundant robots. This is especially true in performance-demanding tasks, as for robots employed on assembly lines in manufacturing industries, repeatedly performing the same activity. Once the assembly line engineer has defined the tool path in the task space, our planning algorithm unifies inverse kinematics and time parametrization so as to bring the manipulator at its physical limits to achieve specific efficiency goals, being execution time the most typical one. The algorithm is configurable in terms of constraints to consider and objective functions to optimize, therefore it can be easily adapted to optimize other custom-defined efficiency indices, to better respond to the needs of the automation plant. Being based on discrete dynamic programming, the global optimum is guaranteed for a given resolution of the problem. This can be configured by the operator to achieve the desired trade-off between efficiency and planning time. In our experiments, we go through the whole process of planning and executing a time-optimal trajectory on a real robot, and discuss some practical details, such as trajectory smoothness and actuator saturation, aiding the practitioners in deploying our algorithm effectively. Currently, the algorithm's applicability is limited to those cases where hours are available for planning, hence it is not well-suited for those cases where the robot activity has to change frequently. By replacing the underlying dynamic programming engine with a different methodology, such as randomized algorithms, the planning time could be controlled to be upper-bounded, thus returning the most efficient solution that can be achieved in the time available for reconfiguring the production. Other applications of interest include optimal ground control of space robotic assets and performance benchmarking of online planning algorithms.
\end{abstract}


\section{Introduction}

\subsection{Scope and Objective}

\IEEEPARstart{K}{inematically} redundant manipulators possess more degrees of freedom than those strictly required to execute a given task. Such a characteristic gives the system a higher degree of dexterity and mobility that can be exploited to optimize performance indices of interest, besides fulfilling the main task. Redundancy can be structural, when the robot is designed to be redundant for generic tasks, or functional, when the robot is operated so as to be redundant. The concept applies to anthropomorphic serial chains, that are by far the most widespread robotic systems in manufacturing industries, but it is general enough to be extended to more complex systems, including humanoids, mobile manipulators (terrestrial, aerial), parallel robots and systems of cooperating robots. The optimization problem that looks into the exploitation of redundancy to optimize a given performance index is commonly referred to as \emph{redundancy resolution}.

\IEEEpubidadjcol

Many robotic tasks require to follow a specific path, such as in welding, cutting, gluing and in some assembly and disassembly tasks. Planning a collision-free task space path is also a convenient solution when the environment is cluttered. This is opposed to \emph{point-to-point} (PTP) motion, where the robot links are more loosely constrained during motion. The definition of the speed by which the path is tracked is also a decision variable that, in most applications, is necessary to optimize. The typical objective is time minimization, but any performance index involving derivatives can be designed for a specific purpose. The optimization problem that looks into the definition of velocity along a given path is commonly referred to as \emph{optimal trajectory planning} or \emph{optimal time-parametrization}.

Both the optimization problems above can be addressed (and solved) at planning level, sometimes offline, before the robot moves, or at control level, while the task is being executed and the system has to react to unforeseen events. Although, nowadays, most applications require the robots to live and operate in unstructured, unknown and highly-dynamical environments, there still are many situations in which tasks are better planned offline. One of the most frequent employments of anthropomorphic arms in manufacturing industries concerns the execution of repetitive tasks in structured environments, where the task is planned once and executed cyclically. In aerospace, offline planning instances exist in the mission design phase for feasibility and budget assessments. Also, some sequences for in-orbit manipulation can be pre-planned in an optimal way. In missions that are characterized by windowed communications, a long planning time is usually available on ground to deliver more efficient commands to the spaceborne asset. On the contrary, when the domain requires highly-reactive behaviors, offline planning can provide benchmark solutions to evaluate the performance of online control algorithms. The scenarios just described are the domain of this paper.

\begin{figure}[!ht]
    \centering   
    \includegraphics[width=0.49\textwidth]{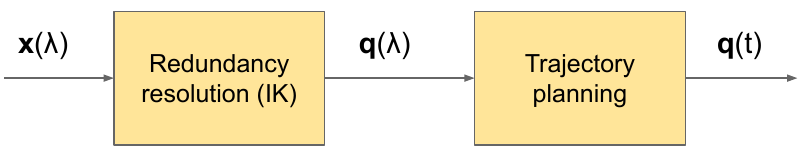}
    \caption{Redundancy resolution and trajectory planning as two independent processes; ${\bf x}(\lambda)$ is the task space path, ${\bf q}(\lambda)$ the joint space path and ${\bf q}(t)$ the joint space trajectory.}    
	\label{fig:rr_totp_scheme}
\end{figure}

Whether they are executed on-line or offline, redundancy resolution and trajectory planning have been mostly treated as two independent, consecutive optimization processes \cite{Chiacchio90, Basile, AlKhudir}, as depicted in Fig.\ \ref{fig:rr_totp_scheme}. A task space path is given, that is mapped, through inverse kinematics, in the joint space (possibly optimizing a geometric quality index), then, the joint space path is time-parametrized to yield a trajectory (respecting constraints and possibly optimizing a different quality index), that can be executed through motion control at joint level. The optimization performed at the former stage can be functional to the optimization performed at the latter. For instance, if the objective of trajectory planning is to define a minimum-time motion, redundancy resolution may optimize, according to a heuristic approach, the acceleration/deceleration capabilities of the manipulator along the path. Overall, the two processes ``cooperate'' to achieve a common objective.

Unfortunately, the resolution of two independent optimal problems does not guarantee the achievement of the overall global optimum. Indeed, if the objective is to optimize a generic dynamic index, e.g., traveling time, but the time law is only defined at the level of trajectory planning, there is no possibility to formulate a redundancy resolution problem which explicitly considers it.

\begin{figure}[!ht]
    \centering   
    \includegraphics[width=0.33\textwidth]{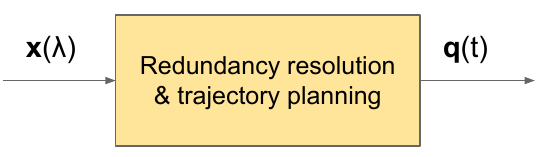}
    \caption{Redundancy resolution and trajectory planning as one unique process; ${\bf x}(\lambda)$ is the task space path and ${\bf q}(t)$ the joint space trajectory.}    
	\label{fig:totpr_scheme}
\end{figure}

Hence, the objective of this paper is to provide a unified solution for both optimization problems, as in Fig.\ \ref{fig:totpr_scheme}, so that kinematic redundancy is effectively exploited for the minimization/maximization of the objective function defined at trajectory planning level.

\subsection{Related works}
\label{sec:related_works}

Optimal trajectory planning along prescribed paths deals with the optimization of an integral cost function, subject to the kinematic constraint of the end-effector's path and dynamic capabilities of the actuators (mainly torque and torque rate limits). The foundations of the problem were laid for time minimization and non-redundant robots \cite{Bobrow, Shin85, Pfeiffer, Slotine}, where the unique joint space path is parametrized with respect to the curvilinear coordinate of the path $\lambda$. The robot kinematic and dynamic constraints are encoded on limits on its first- and second-order time-derivatives $\dot{\lambda}$ and $\ddot{\lambda}$, called \emph{pseudo-velocity} and \emph{pseudo-acceleration}, respectively. In this new domain, $\lambda$ and $\dot{\lambda}$ represent the state variables and $\ddot{\lambda}$ the input of a parametrized dynamic system. The optimal solution is bang-bang in $\ddot{\lambda}$ \cite{Chen92}. If, for a redundant manipulator, the joint space path is already assigned, the theory of non-redundant manipulators holds without modifications.

The first extension to kinematically redundant systems, that includes the more generic case of a path assigned in the task space, was developed in \cite{Galicki99, Galicki00}. Therein, a non-minimal extended state is considered, made of $\lambda$, $\dot{\lambda}$, the joint positions and their derivatives. The input corresponds to actuation torques. The time-optimal problem is solved numerically by making use of the \emph{Extended Pontryagin's Maximum Principle (EPMP)} and considering an approximation of the torque constraints. This formulation is inflexible with respect to the introduction of additional constraints and objective functions.

By adopting a suitable parametrization of redundancy \cite{FerrentinoARK}, e.g.\ through \emph{joint space decomposition (JSD)} \cite{Reiter15}, a minimal state of the parametrized dynamic system can be found, composed of $\lambda$, $\dot{\lambda}$, the redundancy parameter, and its first-order derivative. This time, the equations can be arranged to let the input be the vector made of $\ddot{\lambda}$ and the acceleration of the redundancy parameter, but they cannot be constrained explicitly. As a consequence, a linear programming (LP) problem has to be defined for each possible state in order to determine the optimal input, that is extremized according to the bang-bang paradigm. In \cite{Ma04}, in order to provide an initial state, the initial joint positions are fixed and the resolution is achieved through a try-and-error approach. In addition, in order to make the algorithm more efficient, the number of switching points is possibly reduced with respect to the real time-optimal trajectory. Objective functions other than tracking time cannot be considered, as the bang-bang hypothesis is made. Lastly, the algorithm cannot enforce zero joint velocity at the end of the motion, which is undesirable in most applications. 

A different approach consists in addressing the optimization problem directly \cite{Reiter16a, Reiter16b, Reiter17, Reiter18}, without any prior manipulation of the quantities in play. The time axis is divided into a fixed number of intervals of the same variable length, whose sum is $t_f$, that is the result of the optimization process. When $t_f$ is given, the duration of each time interval in the time domain is also determined. The problem is to find, for each of the intervals, the value of $\lambda$, the joint position vector, and their derivatives. The path constraint is imposed through second-order kinematics, either with JSD or null-space projection. For both techniques, since the path constraint is enforced through numerical integration of the second-order solution, an additional positional kinematics constraint needs to be included to prevent drift. The problem can be solved with direct multiple shooting using the interior-point method, after an initial guess on the optimizing variables is made. Since the problem is non-convex, globally-optimal solutions cannot be guaranteed and, in general, the quality of the resulting solution strictly depends on the initial guess. On the other hand, the formulation is flexible enough to include constraints on the joint positions and their derivatives, as well as other application-specific constraints. For instance, in \cite{Reiter18}, the path constraint is further derived to account for the limits on joint positions, velocities, accelerations, jerks and snaps (i.e.\ the fourth-order derivative of the position) through which continuous and quasi-differentiable motor torques can be obtained. By acting directly on $\ddot{\lambda}$, this is also the objective in \cite{Wang23}. This result is of utmost importance since the insufficient continuity of globally-optimal minimum time trajectories is the main shortcoming when it comes to the control of real robots.

For the sake of completeness, we note that a multitude of different optimization approaches exist in more recent literature on trajectory planning. They include dynamic optimization in the form of convex quadratic \cite{Wolinski, Zhang23} or cone \cite{Marauli} programming, evolutionary algorithms \cite{Ogbemhe, Wang22, Zhang21, Wang21}, dynamic programming \cite{Schappler}, heuristic algorithms \cite{Singh22}, and machine learning \cite{Ma23}, with several approaches to describe the solution, in the form of, e.g., polynomials \cite{Chen21, Singh22} and B-splines \cite{Ogbemhe, Wang22, Zhang21, Wang21} to control the joint space smoothness or achieve multi-objective optimization \cite{Ogbemhe, Zhang23, Wu23, Wang22, Zhang21, Wang21}. The mentioned techniques, however, are only applied to solve sub-problems of the unified optimal redundancy resolution and time parametrization problem addressed here. They consider, for instance, PTP instead of path-constrained tasks \cite{Wang22, Wang21}, or only the first \cite{Wolinski, Zlajpah} or second \cite{Marauli, Chen21, Wu23} stage in Fig.\ \ref{fig:rr_totp_scheme}.

\subsection{Methodology and contribution}

In all the offline planning applications that motivate this work, although the environment where the robot operates is structured, static, known or partially known, not all the formulations of the optimization problem can capture the complexity of the real system and some show greater flexibility than others. This paper adopts discrete dynamic programming (DDP) as the main underlying methodology and central idea to cope with the complexity of reality. Also, because of its employment, in previous works, for both the optimization problems described above, it naturally arises as the unifying approach (see, for instance, \cite{Pashkevich, Dolgui, Guigue10, Gao, FerrentinoJMR} for redundancy resolution, and \cite{Shin86, Singh87, Kaserer19, Kaserer20} for trajectory planning). DDP is extremely flexible in the accommodation of arbitrary constraints and objective functions and, compatibly with the available planning time, can achieve the global optimum. When applied to real systems, its formulation can well describe the constraints at hand and no conservative hypotheses are necessary. The computational complexity of DDP is square with the cardinality of the discrete sets defining the feasible search space \cite{FerrentinoJMR}. However, the more constraints are defined in the problem formulation (which is the typical case of real systems), the more states are excluded from the search. Therefore, differently from several other numerical optimization methods, e.g., those based on the Newton's method, where more constraints make the search of the optimum cumbersome, in DDP they are beneficial to reducing the computation time.

DDP is based on the discretization of the state space which, in principle, complicates the generation of smooth references for motion control. However, the possibility to include higher-order constraints mitigates this issue.

Our contribution is a unified generic framework for redundancy resolution and trajectory planning based on DDP. With respect to previous works proposing a unified solution, i.e.\ \cite{Reiter15, Reiter16a, Reiter16b, Reiter17, Reiter18}, our methodology does not need the user to provide a seed state, i.e., an initial state from which the search starts, and which closely influences the final solution, but can retrieve the globally-optimal solution of the discretized state space. Possibly, this solution can be used as a valuable seed state for a different optimization problem to further improve other features of the solution, such as its smoothness. We further elaborate on this aspect in Section \ref{sec:additional_trials}.

\subsection{Overview}

In Section \ref{sec:problem_formulation}, the problem is presented and formalized. First, parameters and equations that describe the problem are introduced in Section \ref{sec:state_definition} and discretized so as to identify a suitable state that provides a minimal description. Then, in Section \ref{sec:constraints}, the constraints are introduced as maps that enable transitions towards certain states, in a generic framework. Last, in Section \ref{sec:dp_formulation}, the optimization problem is introduced in a generic way and specialized for time-optimal planning. The algorithmic formulation is provided in Section \ref{sec:algorithm}, together with some important considerations of practical interest. The results of the experiments are presented in Section \ref{sec:experiments}. First, the experimental setup is described in Section \ref{sec:experimental_setup}, then, results of offline planning and execution on a real robot are given in Section \ref{sec:planning} and Section \ref{sec:execution} respectively. In Sections \ref{sec:comparison_two_stage} and \ref{sec:comparison_ms}, our approach is compared with two algorithms from \cite{Chiacchio90} and \cite{Reiter16a, Reiter16b, Reiter17, Reiter18}, using the decoupled approach of Fig.\ \ref{fig:rr_totp_scheme} and another unified approach as in Fig.\ \ref{fig:totpr_scheme}, respectively. The work is concluded in Section \ref{sec:conclusion}, where some lines of development are also outlined for future research.

\section{Optimal planning of redundant robots with dynamic programming}
\label{sec:problem_formulation}

\subsection{State definition}
\label{sec:state_definition}

Let us consider a geometrical path for the end-effector ${\bf x}(\lambda):[0,L] \to \mathbb{R}^m$, such that a given manipulator with $n$ degrees of freedom is redundant, i.e.\ $n>m$. It follows that $r = n - m$, the \emph{redundancy degree}, is strictly greater than zero. $\lambda$ is the arc length, $L$ is the path length. The joint positions along the path ${\bf q}(\lambda) \in \mathbb{R}^n$ are unknown, as well as the time law $\lambda(t):\mathbb{R} \to [0,L]$ associated with them. The time derivative of the arc length $\dot{\lambda} = \frac{d \lambda}{dt}$ encodes the time information, and, when expressed as $\dot{\lambda}(\lambda)$, for variable $\lambda$, it is often referred to as \emph{phase plane trajectory} (PPT) \cite{Bobrow, Shin85}. Intuitively, the PPT corresponds to the norm of the tangential velocity with respect to the arc length.

Let us discretize $\lambda$ in space with a given step $\Delta_{\lambda}$ and $N_i \in \mathbb{N}$, i.e.,
\begin{equation} \label{lambda_sampling}
{\lambda}(i) = i\Delta_{\lambda}, \quad \mbox{with}~i=0,1,2,\ldots,N_i \quad \mbox{and}~ \Delta_{\lambda} = \frac{L}{N_i}
\end{equation}
such that ${\bf x}\big(\lambda(i)\big) = {\bf x}(i)$ can be viewed as the $(i+1)$-th waypoint. The variable $i$ is termed \emph{stage index}.

The robot dynamic model is given by
\begin{equation} \label{torque_sample}
\boldsymbol{\tau}(i) = {\bf H}\big( {\bf q}(i) \big) \ddot{\bf q}(i) + {\bf f} \big( {\bf q}(i), \dot{\bf q}(i) \big),
\end{equation}
where, by omitting the dependence on $i$ for simplicity, ${\bf H}({\bf q})$ is the $n \times n$ inertia matrix and
\begin{equation} \label{f_vector}
{\bf f}({\bf q}, \dot{\bf q}) = \dot{\bf q}^T{\bf C}({\bf q})\dot{\bf q} + {\bf d}(\dot{\bf q}) + {\bf g}({\bf q}).
\end{equation}
${\bf C}({\bf q})$ is the $n \times n \times n$ matrix of Christoffel symbols, ${\bf d}(\dot{\bf q})$ is the $n \times 1$ vector of Coulomb and viscous friction terms, and ${\bf g}({\bf q})$ is the $n \times 1$ vector of gravitational torques.

Let $S_{(\cdot)}(i)$ be the discrete-time function (or \emph{sequence} of samples) of the continuous-time function $(\cdot)(\lambda)$ up until the $i$-th sample. For instance, $S_{\dot{\lambda}}(i) = \lbrace \dot{\lambda}(0),\,\ldots,\,\dot{\lambda}(i) \rbrace$ is the discrete-time phase plane trajectory up until stage $i$.

The joint velocities and accelerations at a given stage $i$, $\dot{\bf q}(i)$ and $\ddot{\bf q}(i)$, respectively, in a discrete-time formulation, can be approximated, through fixed-step integration schemes, as functions of the sequences $S_{\bf q}(i)$, $S_{\dot{\bf q}}(i)$ and $S_t(i)$, i.e.,
\begin{align}
\dot{\bf q}(i) & \sim f_{\dot{\bf q}}\big( S_{\bf q}(i), S_t(i) \big) \label{joint_velocity_sample}, \\
\ddot{\bf q}(i) & \sim f_{\ddot{\bf q}}\big( S_{\dot{\bf q}}(i), S_t(i) \big). \label{joint_acceleration_sample}
\end{align}
The exact form of the functions above depends on the specific integration scheme.  We note here, and further clarify in Section \ref{sec:algorithm}, that derivatives are assumed to be computed with past samples, in a backward integration scheme, because future samples, in a forward optimization scheme, are unknown, as they depend on optimal choices that yet have to be made.

In the same way, each timestamp $t(i)$ in $S_t(i)$ can be derived from the phase plane trajectory up until stage $i$, through some discrete approximation, e.g.\ the trapezoidal approximation used in \cite{Singh87}, that generically is
\begin{equation} \label{time_sample_discrete}
t(i) \sim f_t \big( S_{\lambda}(i), S_{\dot{\lambda}}(i) \big).
\end{equation}
Let us remark, once again, that $S_{\dot{\lambda}}(i)$ is unknown and, as a consequence, $t(i)$ is unknown, too.

Let us define the \emph{forward kinematics function} ${\bf k}({\bf q}): \mathbb{R}^n \to \mathbb{R}^m$. The path constraint at the given waypoints is given by
\begin{equation} \label{dk}
{\bf x}(i) = {\bf k}\big( {\bf q}(i) \big).
\end{equation}
In order to parametrize the redundancy, we may adopt a joint selection (or JSD) or a joint combination scheme \cite{FerrentinoARK}, i.e.\ the redundancy parameters are selected as a subset of the joint positions or they are a more generic function of them. Under this hypothesis, each of the joint positions in $S_{\bf q}(i)$ in \eqref{joint_velocity_sample} can be computed from non-redundant inverse kinematics (IK) as
\begin{equation} \label{non-redundant_ik}
{\bf q}(i) = {\bf k}^{-1} \big({\bf x}(i), {\bf v}(i), g(i) \big),
\end{equation}
where ${\bf v} \in \mathbb{R}^r$ is the vector of redundancy parameters. Since more than one configuration may satisfy the generic non-redundant IK, $g$ is a selector of the specific IK solution to be considered at stage $i$. For example, if the square inverse kinematics function ${\bf k}^{-1}({\bf x}, {\bf v})$ admits 4 solutions, $g \in \lbrace 0, 1, 2, 3 \rbrace$. This way defined, ${\bf q}(i)$ is unique. Since ${\bf x}$ is a function of $\lambda$, we may rewrite the equation above as
\begin{equation} \label{joint_position_sample}
{\bf q}(i) = {\bf k}^{-1} \big(\lambda(i), {\bf v}(i), g(i) \big).
\end{equation}

Now, we may fold \eqref{time_sample_discrete} and \eqref{joint_position_sample}, for each $i$, into \eqref{joint_velocity_sample} and \eqref{joint_acceleration_sample}, yielding
\begin{align}
\dot{\bf q}(i) & \sim f_{\dot{\bf q}}\big( S_{\lambda}(i), S_{\dot{\lambda}}(i), S_{\bf v}(i), S_g(i) \big) \label{totpr_qd_computation}, \\
\ddot{\bf q}(i) & \sim f_{\ddot{\bf q}}\big( S_{\lambda}(i), S_{\dot{\lambda}}(i), S_{\bf v}(i), S_g(i) \big) \label{totpr_qdd_computation}.
\end{align}

By substituting both equations above into \eqref{torque_sample}, we find that, in general, the input is a function of the four identified sequences, i.e.\
\begin{equation} \label{totpr_torque_computation}
\boldsymbol{\tau}(i) \sim f_{\boldsymbol{\tau}} \big( S_{\lambda}(i), S_{\dot{\lambda}}(i), S_{\bf v}(i), S_g(i) \big).
\end{equation}

This means that, given the current state at stage $i$, selecting the input $\boldsymbol{\tau}(i)$ that drives the discrete-time system to another state is equivalent to selecting the parameters $\dot{\lambda}$, ${\bf v}$ and $g$ at the next stage. This is coherent with the logic of discretizing the state space instead of the input space that is common to all the previous works using DDP for redundancy resolution and time-optimal planning. Notoriously, this is also a way to control the number of involved variables due to the addition of extra dimensions, i.e.\ the \emph{curse of dimensionality} \cite{Bellman}, that is a well-known issue in dynamic programming. The parameters $\dot{\lambda}$, ${\bf v}$ and $g$ also constitute a minimum representation of the state for each stage, as opposed to \cite{Galicki99} and \cite{Reiter16a} that, rather, use redundant representations. In a formulation based on calculus of variations, this result is related to the minimum number of differential equations required to fully represent the system at hand \cite{FerrentinoARK}.

Thus, let us proceed by discretizing the variables $\dot{\lambda}$ and ${\bf v}$. The former is
\begin{equation} \label{totpr_lambdad_discretization}
\dot{\lambda}_l = l\Delta_{\dot{\lambda}}, \quad \mbox{with}~ l=0,1,2,\ldots,N_l\quad \mbox{and}~ \Delta_{\dot{\lambda}} = \frac{\dot{\lambda}_M}{N_l},
\end{equation}
where $\dot{\lambda}_M$ is the maximum pseudo-velocity value that the phase plane trajectory can reach. The latter is
\begin{equation} \label{totpr_rp_discretization}
{\bf v} = {\bf j} \circ {\bf \Delta}_v + {\bf v}_{min},
\end{equation}
where `$\circ$' denotes the element-wise Hadamard product, ${\bf \Delta}_v = [\Delta_{v,1}, \ldots, \Delta_{v,r}]$ is the vector of the sampling intervals, ${\bf j} \in \mathbb{N}^r$ the vector of the redundancy parameter indices, and ${\bf v}_{min}$ is the vector of lower bounds of the redundancy parameters domains. The elements of ${\bf j}$ take the maximum values $N_{j,1}, \ldots, N_{j,r}$ so that $[N_{j,1}, \ldots, N_{j,r}] \circ {\bf \Delta}_v + {\bf v}_{min}$ equals the upper bound of the redundant joint domains.

The discretized domains of $\lambda$, $\dot{\lambda}$, ${\bf v}$ and $g$ (which is naturally discrete) makes up a grid of $r+3$ dimensions. We essentially adopt the same approach as other DDP formulations for redundancy resolution and time-optimal planning, but we increase the number of dimensions to take into account: (1) the resolution of both problems together; (2) the presence of multiple inverse kinematics solutions.

Let us recall that, for non-redundant manipulators, the curvilinear coordinate parametrization of dynamics is obtained through the following equations:
\begin{equation} \label{nr_path_parametrization}
\begin{split}
\dot{\mathbf{q}} = & {\bf J}^{-1}{\bf x}' \dot{\lambda}, \\
\ddot{\mathbf{q}} = & {\bf J}^{-1} \left[ \mathbf{x}' \ddot{\lambda} + \mathbf{x}'' \dot{\lambda}^2 - ({\bf J}^{-1} \mathbf{x}')^T {\bf B}({\bf J}^{-1} \mathbf{x}') \dot{\lambda}^2 \right],
\end{split}
\end{equation}
where ${\bf J} = \frac{\partial {\bf k}}{\partial {\bf q}}$ is the manipulator's Jacobian matrix, ${\bf x}' = \frac{d{\bf x}}{d\lambda}$, ${\bf x}'' = \frac{d^2{\bf x}}{d\lambda^2}$, and ${\bf B}$ is the Hessian of the vector function ${\bf k}$ \cite{Slotine}. Equations \eqref{nr_path_parametrization} are substituted in \eqref{torque_sample} to yield the parametrized dynamics. In the redundant case considered here, ${\bf J}$ is no longer invertible, compromising the possibility of a unique parametrization. As a consequence, the pseudo-acceleration limits cannot be pre-computed: a state space grid may still be computed on the basis of the discrete values of the state, but the content of each single node should be reconsidered in light of this observation. Each node in the grid will then contain a state given by 
\begin{equation} \label{totpr_node_content}
{\bf s}_{l{\bf j}g}(i) = [\dot{\lambda}_l(i), {\bf q}_{{\bf j}g}(i)],
\end{equation}
where ${\bf q}_{{\bf j}g}$ is the vector of joint positions obtained from \eqref{joint_position_sample} using the redundancy parameters ${\bf v_j}$ and selecting the $g$-th IK solution.

In what follows, for the sake of simplicity, whenever we do not need to refer to the specific position that a given state occupies in the grid, we drop the state's subscripts and refer to the generic state at stage $i$ as ${\bf s}(i)$.

\subsection{Constraints}
\label{sec:constraints}

In offline robotic planning, common constraints are limits on joint positions and their derivatives, actuation constraints, obstacles in the workspace, velocity and acceleration limits in the Cartesian space, initial/final joint configurations, cyclicity (for closed task space paths), power limits, maximum/minimum forces exchanged with the environment in an interaction scenario. Many other constraints can be designed depending on the specific application at hand. In the experiments that will follow in Section \ref{sec:experiments}, we consider minimum/maximum joint positions, maximum joint velocities, accelerations, jerks, torques and torque rates \cite{Petrone}.

Since the robot dynamics cannot be parametrized with respect to $\lambda$ and its derivatives, the same holds for the constraints. Therefore, for redundant robots, they are directly verified on the joint variables. Hence, let us define the stage-dependent set $\mathcal{A}_i$, containing all nodes returning joint positions that respect the joint domains, the path constraint and, possibly, imposed initial/final configurations. Such constraints are formalized as
\begin{equation}
{\bf s}(i) \in \mathcal{A}_i.
\end{equation}
Joint velocity, acceleration and jerk limits can be directly encoded in equivalent stage-dependent, as well as state-dependent sets $\mathcal{B}^1_i$ $\mathcal{B}^2_i$ and $\mathcal{B}^3_i$, so that:
\begin{align} 
\dot{\bf q}(i) & \in \mathcal{B}^1_i\big({\bf s}(i)\big) \label{totpr_velocity_constraint}, \\
\ddot{\bf q}(i) & \in \mathcal{B}^2_i\big({\bf s}(i)\big) \label{totpr_acceleration_constraint}, \\
\dddot{\bf q}(i) & \in \mathcal{B}^3_i\big({\bf s}(i)\big) \label{totpr_jerk_constraint}.
\end{align}
Equivalently, joint torque and torque rate limits can be encoded in similar sets $\mathcal{C}^1_i$ and $\mathcal{C}^2_i$:
\begin{align}
\boldsymbol{\tau}(i) & \in \mathcal{C}^1_i\big({\bf s}(i)\big) \label{totpr_torque_constraint}, \\
\dot{\boldsymbol{\tau}}(i) & \in \mathcal{C}^2_i\big({\bf s}(i)\big) \label{totpr_torque_rate_constraint}.
\end{align}
More commonly, all the quantities above are given within fixed domains that do not change along the trajectory, are not configuration dependent, and not velocity-dependent, but the generality of the framework allows for the accommodation of more complex constraints. For example, interacting torque constraints expressed as
\begin{equation} \label{interacting_torques_constraint_generic}
\boldsymbol{\tau} = [\tau_1, \tau_2, \ldots, \tau_n]^T \in E(\mathbf{q}, \dot{\mathbf{q}}),
\end{equation}
that cannot be parametrized \cite{Shin86}, can be easily handled with DDP.

All the sets above can be combined together to define the set $\mathcal{D}_i$  of reachable states for a generic stage $i$, that is
\begin{equation} \label{all_constraints}
\begin{split}
\mathcal{D}_i = \mathcal{A}_i \cap \Bigg\{ {\bf s}(i): \dot{\bf q}(i) \in \mathcal{B}^1_i, \ddot{\bf q}(i) \in \mathcal{B}^2_i, \dddot{\bf q}(i) \in \mathcal{B}^3_i, \\
\boldsymbol{\tau}(i) \in \mathcal{C}^1_i, \dot{\boldsymbol{\tau}}(i) \in \mathcal{C}^2_i  \\
\mbox{with}~ {\bf s}(i-1) \in \mathcal{A}_{i-1}, \ldots,{\bf s}(0) \in  \mathcal{A}_{0} \Bigg\}.
\end{split}
\end{equation}

\subsection{Discrete dynamic programming problem}
\label{sec:dp_formulation}

Since all the states are available in the grid, the DDP problem is transformed into a graph search problem. This allows to adopt either a forward or a backward optimization scheme \cite{FerrentinoJMR}. The objective function to optimize can be generally defined as
\begin{equation} \label{dp_setting}
I(N_i) = \psi \big( {\bf s}(0) \big) + \sum^{N_i}_{k = 1} \phi \big( \mathbf{s}(k-1), \mathbf{s}(k) \big),
\end{equation}
where $\psi \big( {\bf s}(0) \big)$ is a cost associated to the initial state and $\phi$ is the cost computed locally between two adjacent states. 

If, for instance, the objective function to minimize is the time, there is no cost associated to the initial state, i.e.\ $\psi \big( {\bf s}_0 \big) = 0$ and $I=t$. Assuming a backward Euler approximation, \eqref{dp_setting} becomes
\begin{equation} \label{dp_setting_time}
t(N_i) = \sum_{k=1}^{N_i} \frac{\lambda(k) - \lambda(k-1)}{\dot{\lambda}(k)}
\end{equation}
A different approximation can be adopted at the beginning and at the end of the trajectory to cope with the case $\dot{\lambda}=0$ \cite{Singh87}.

To complete our formulation, we rewrite \eqref{dp_setting_time} in a recursive form and use the set $\mathcal{D}_i$ in \eqref{all_constraints}, by minimizing over the admissible states:
\begin{equation} \label{totpr_optimal_return_function}
\begin{split}
t_{opt}(0) & = 0, \\
t_{opt}(i) & = \min_{{\bf s}(i) \in \mathcal{D}_i} \left[ t_{opt}(i-1) + \frac{\lambda(i) - \lambda(i-1)}{\dot{\lambda}(i)} \right],
\end{split}
\end{equation}
where $t_{opt}(i)$ at a generic stage $i$ is the \emph{optimal return function} (or \emph{value function}) and $t_{opt}(N_i)$ represents the optimal cost.

\section{Algorithmic implementation}
\label{sec:algorithm}

Based on the problem formulation above, we can define the time-optimal trajectory planning algorithm for redundant robots (TOTP-R) with DDP as in Algorithm \ref{alg:totpr_dp_algo}.

\begin{algorithm}
\centering
\small
\begin{algorithmic}[1]
\State \emph{Initialize state space grid through inverse kinematics and discretization of $\dot{\lambda}$, according to \eqref{joint_position_sample} and \eqref{totpr_node_content}} \label{stp:grid_initialization}
\State \emph{Initialize ${\cal A}_i, \forall i=0..N_i$} \label{stp:initialize_Ai}
\State \emph{Initialize ${\cal B}_i^1, {\cal B}_i^2, {\cal B}_i^3, {\cal C}_i^1, {\cal C}_i^2 ~\forall i=0..(N_i-1)$ with state-independent information} \label{stp:initialize_Bi}
\State \emph{Initialize ${\cal D}_i = \varnothing, \forall i=0..N_i$}
\State \emph{Initialize cost map $t_{i,l,{\bf j},g} = +\infty ~\forall i,l,{\bf j},g$} 
\State $t_{0,l,{\bf j},g} \gets 0 ~\forall l,{\bf j},g$
\State ${\cal D}_0 \gets {\cal A}_0$
\For{$i \gets 0$ to $N_i-1$}
\For{\textbf{each} ${\bf s}_{l{\bf j}g} \in {\cal D}_{i}$} \label{stp:left_node}
\For{\textbf{each} ${\bf s}_{m{\bf k}h} \in {\cal A}_{i+1}$} \label{stp:right_node}
\State \emph{Compute $\dot{\bf q}$, $\ddot{\bf q}$, $\dddot{\bf q}$, $\boldsymbol{\tau}$, $\dot{\boldsymbol{\tau}}$}
\If{Constraints in \eqref{totpr_velocity_constraint} - \eqref{totpr_torque_rate_constraint} are satisfied} \label{stp:constraints_check}
\State ${\cal D}_{i+1} \gets {\cal D}_{i+1} \cup \{{\bf s}_{m{\bf k}h}\}$ \label{stp:node_enabling}
\State \emph{Compute instantaneous cost function $\phi$}
\If{$t_{i,l,{\bf j},g} + \phi < t_{i+1,m,{\bf k},h}$}
\State $t_{i+1,m,{\bf k},h} \gets t_{i,l,{\bf j},g} + \phi$ 
\State \emph{Let ${\bf s}_{l{\bf j}g}$ at stage $i$ be the predecessor of ${\bf s}_{m{\bf k}h}$ at stage $i+1$} \label{stp:update_pred_map} 
\EndIf
\EndIf
\EndFor
\EndFor
\EndFor
\State $t_{opt}(N_i) \gets \min_{{\bf j},g} \left[ t_{N_i,0,{\bf j},g} \right]$ \label{stp:minimum_time}
\State \emph{Build functions $\dot{\lambda}(i)$ and ${\bf q}(i)$ of optimal pseudo-velocities and joint positions by screening the predecessors map backward} \label{stp:trajectory_generation}
\end{algorithmic}
\caption{DDP TOTP-R algorithm.}
\label{alg:totpr_dp_algo}
\end{algorithm}

The algorithm assumes that a task space path is given as a discrete set of points and that they are associated to discrete values of the curvilinear coordinate, from $\lambda = 0$ up to the length of the path $\lambda = L$. If this is not the case, but some analytic curve is given, it can be sampled at the desired rate. Although \eqref{lambda_sampling} is a uniform sampling, this is not a necessary condition for TOTP-R. At step \ref{stp:grid_initialization}, the redundancy parameters vector is discretized according to \eqref{totpr_rp_discretization}, which allows to compute IK solutions as in \eqref{joint_position_sample}. The pseudo-velocity is also discretized according to \eqref{totpr_lambdad_discretization}, so that the grid nodes can be defined as in \eqref{totpr_node_content}.

In agreement with \cite{Pamanes, FerrentinoJMR}, let us define the \emph{extended aspect} as each connected set of joint configurations ${\bf q}$, such that the determinant of the augmented (with the redundancy parameters) kinematics' Jacobian $\lvert {\bf J}_a({\bf q}) \rvert$ never nullifies. On the basis of this definition, every combination of signs of the factors of $\lvert {\bf J}_a({\bf q}) \rvert$ identifies a different extended aspect \cite{Wenger}, achieving a partition of the configuration space. If these combinations are known to the IK solver in \eqref{joint_position_sample}, IK solutions belonging to different extended aspects can be mapped to different grids. In this case, $g$ is also an extended aspect selector and, if desired, can be fixed in \eqref{joint_position_sample} to avoid traversing singularities pertaining to the augmented Jacobian. However, in general, finding a proper factorization of $\lvert {\bf J}_a({\bf q}) \rvert$ is not trivial, in which case the aforementioned mapping cannot be established and the state space grids obtained through \eqref{joint_position_sample} are \emph{non-homogeneous} \cite{FerrentinoJMR}, i.e.\ each grid possibly mixes solutions from different extended aspects. As a consequence, $g$ cannot be used for limiting the manipulator's motion to specific sub-spaces of the configuration space.

At step \ref{stp:initialize_Ai}, the nodes in the grid are enabled/disabled according to geometrical constraints including, for example, configurations that bring the robot to collide with the surrounding environment. Also, if IK solutions are homogeneous, entire grids can be excluded to make the DDP algorithm find a trajectory in specific extended aspects. The sets $\mathcal{A}_i$ can also be used to impose that the robot must start and finish its motion at rest, or even stop along the path if this is required by the specific task. In time-optimal planning of non-redundant robots, the same sets $\mathcal{A}_i$ can be used to impose the maximum velocity curve (MVC) constraint \cite{Bobrow, Shin85}. In this case, since we do not perform any parametrization of dynamics and constraints, this is not possible, and the MVC constraint is implicitly checked at the time of verifying \eqref{totpr_velocity_constraint}-\eqref{totpr_torque_rate_constraint} at step \ref{stp:constraints_check}.

In most practical cases, constraints on joint velocities, accelerations, jerks, torques, and torque rates are given in terms of state-independent connected sets, e.g. $\dot{\bf q} \in \left[ \dot{\bf q}_{min}, \dot{\bf q}_{max} \right]$. In this case, the sets in \eqref{totpr_velocity_constraint}-\eqref{totpr_torque_rate_constraint} are completely defined beforehand, as done at step \ref{stp:initialize_Ai}. Conversely, if they were state-dependent, they could be re-computed at the time the constraints are checked, where the current velocity and acceleration of the system are known.

At a given stage $i$, for each pair of nodes (steps \ref{stp:left_node} and \ref{stp:right_node}) the discrete-time functions (or sequences) of parameters $S_{\lambda}(i)$, $S_{\dot{\lambda}}(i)$, $S_{\bf v}(i)$, $S_g(i)$ are available through back-pointers to compute joint velocities, accelerations and torques as in \eqref{totpr_qd_computation}-\eqref{totpr_torque_computation} respectively, as well as joint jerks and torque rates with equivalent discrete approximations. If constraints are satisfied, we say that the node ${\bf s}_{m{\bf k}h}(i+1)$ can be reached by the node ${\bf s}_{l{\bf j}g}(i)$. The former is then added to the set of nodes $\mathcal{D}_{i+1}$ that can be visited at the next stage (step \ref{stp:node_enabling}). Among all the nodes at stage $i$ that can reach ${\bf s}_{m{\bf k}h}(i+1)$, at step \ref{stp:update_pred_map}, we save the pointer to the one that provides the lowest cumulative cost (or optimal return function), according to \eqref{totpr_optimal_return_function}.

After the process above has been repeated for all the task space points, the one with minimum cumulative traversing time is picked (step \ref{stp:minimum_time}) and, from it, the information about the time law, i.e.\ $\dot{\lambda}(i)$, and exploitation of the null-space, i.e.\ $\dot{\bf q}(i)$, are retrieved by following the map of predecessors backwards (step \ref{stp:trajectory_generation}). The timestamps can be computed through \eqref{time_sample_discrete} and applied to the joint space path to obtain the optimal joint space trajectory. Likewise, optimal joint velocities, accelerations, and torques can be computed from \eqref{totpr_qd_computation}-\eqref{totpr_torque_computation}.

A few considerations are worthwhile. The enforcement of higher-order derivative constraints, like \eqref{totpr_jerk_constraint} and \eqref{totpr_torque_rate_constraint}, would in principle require the introduction of an additional axis corresponding to the pseudo-acceleration $\ddot{\lambda}$ \cite{Shin86, Pham17}. The employment of jerk and/or torque rate constraints becomes particularly important when the trajectories have to be executed on the real hardware, however the introduction of an additional dimension might make the problem intractable from the practical standpoint. This is the issue addressed in \cite{Kaserer19}, where higher-order constraints are also enforced in a planning scenario based on dynamic programming, without redundancy. Therein, continuous and differentiable profiles are generated in the phase plane through interpolation between two consecutive stages. The constraints of interest, including jerk and torque rate, are verified at selected check points along such profiles. This way, the search space does not need to be augmented and the complexity can be controlled without exiting the phase plane. In this paper, we essentially adopt the same solution, but the type of interpolation is left as a user choice as an additional mean to control the complexity. In both cases, when higher-order constraints are checked without augmenting the search space, the global optimality might be compromised \cite{Kaserer19}.

Furthermore, with respect to pure redundancy resolution at kinematic level and pure time-optimal planning for non-redundant robots, the TOTP-R search space is of an increased dimension. Practically, this rules out the usage of an arbitrarily fine discretization and more trade-offs should be considered. Therefore, our solutions might be, in some cases, far from the true global optimum, and we should rely on a more relaxed condition of \emph{resolution optimality}. A proof of convergence of approximate dynamic programming, for the more generic case of multi-objective optimization, is provided in \cite{Guigue09}.

\section{Experimental results}
\label{sec:experiments}

\subsection{Experimental setup}
\label{sec:experimental_setup}

The TOTP-R algorithm is validated on the Panda 7-DOF manipulator by Franka Emika\footnote{https://www.franka.de/panda}, whose joint limits are reported in Table \ref{tab:panda_joint_limits}\footnote{https://frankaemika.github.io/docs/}.

\begin{table}
\centering
\scriptsize
\renewcommand{\arraystretch}{1.2}
\caption{Panda limits from joint 1 (top) to joint 7 (bottom)}
\label{tab:panda_joint_limits}
\begin{tabular}{ccccccc}
$q_{min}$ & $q_{max}$ & $\dot{q}_{max}$ & $\ddot{q}_{max}$ & $\dddot{q}_{max}$ & $\tau_{max}$ & $\dot{\tau}_{max}$ \\
{[rad]} & [rad] & [rad/s] & [rad/s\textsuperscript{2}] & [rad/s\textsuperscript{3}] & [Nm] & [Nm/s] \\
\hline
$-2.8973$ & $2.8973$ & $2.1750$ & $15$ & $7500$ & $87$ & $1000$  \\
$-1.7628$ & $1.7628$ & $2.1750$ & $7.5$ & $3750$ & $87$ & $1000$ \\
$-2.8973$ & $2.8973$ & $2.1750$ & $10$ & $5000$ & $87$ & $1000$ \\
$-3.0718$ & $-0.0698$ & $2.1750$ & $12.5$ & $6250$ & $87$ & $1000$ \\
$-2.8973$ & $2.8973$ & $2.6100$ & $15$ & $7500$ & $12$ & $1000$ \\
$-0.0175$ & $3.7525$ & $2.6100$ & $20$ & $10000$ & $12$ & $1000$ \\
$-2.8973$ & $2.8973$ & $2.6100$ & $20$ & $10000$ & $12$ & $1000$ \\
\hline
\end{tabular}
\end{table}

We rely on the dynamic model identified in \cite{Gaz} and related libraries. They allow access to the inertia matrix ${\bf H}({\bf q})$, the Coriolis vector $\dot{\bf q}^T{\bf C}({\bf q})\dot{\bf q}$, the friction vector ${\bf d}(\dot{\bf q})$, and the gravity vector ${\bf g}({\bf q})$ in \eqref{torque_sample} and \eqref{f_vector}. The friction model includes Coulomb terms and adopts a sigmoidal formulation to avoid discontinuities for low joint velocities.

The execution of the TOTP-R algorithm requires a set of parameters to be defined, that are crucial for the performance of the algorithm, both in terms of quality of the joint space solution and CPU planning time. Indeed, this is common to other techniques (see, for example, \cite{AlKhudir}). They are:
\begin{itemize}
\item number of waypoints in the assigned task space path;
\item pseudo-velocity limit to be used in the grid computation algorithm, which in principle should be large enough to contain the maximum pseudo-velocity that the phase plane trajectory can reach, that is not known beforehand;
\item pseudo-velocity resolution, determining, on one side, the accuracy in the definition of the phase plane trajectory and, on the opposite, the computational complexity of the DDP algorithm;
\item redundancy parameter resolution, determining, on one side, the capability of the redundant manipulator to exploit its null-space to optimize the performance index at hand and, on the opposite, such as before, the computational complexity of the algorithm.
\end{itemize}

The redundancy parameter is selected as the joint position $q_4$ (middle of the chain), because it simplifies the analytical inverse kinematics in \eqref{non-redundant_ik} \cite{Diankov}. As far as the discrete approximations for derivatives are concerned, in our experiments, that are based on a single-threaded implementation, we select a simple backward Euler approximation so as to speed up the computation and consume less memory. However, with a high-performance implementation, more complex discrete approximations could be used and more accurate results should be expected.

\subsection{Planning}
\label{sec:planning}

Two paths are designed in the task space. One is a straight line path having $x = 0.5$ m, $z = 0.4$ m and $y$ spanning for $0.5$ m from $y_s = 0.25$ m to $y_e = -0.25$ m. The other is an ellipse-like line with $L = 1.45$, obtained through cubic splining of the control points (CP) in Table \ref{tab:ellipse_ctrl_points}. The end-effector orientation can be easily understood from the accompanying video\footnote{\url{https://t.ly/ZKelW}}.

\begin{table}
\centering
\scriptsize
\renewcommand{\arraystretch}{1.2}
\caption{Cubic spline control points for the ellipse-like path}
\label{tab:ellipse_ctrl_points}
\begin{tabular}{cccccc}
Coordinate & & & & &\\
in meters [m] & CP\textsubscript{1} & CP\textsubscript{2} & CP\textsubscript{3} & CP\textsubscript{4} & CP\textsubscript{5} \\
\hline
$x$ & $0.5$ & $0.5$ & $0.5$ & $0.5$ & $0.5$ \\
$y$ & $0.0$ & $-0.3$ & $0.0$ & $0.3$ & $0.0$ \\
$z$ & $0.8$ & $0.6$ & $0.5$ & $0.6$ & $0.8$ \\
\hline
\end{tabular}
\end{table}

Both tasks constrain six dimensions, leaving one degree of freedom for redundancy resolution. In addition, the robot has to track the path as fast as possible and is free to select the IK solutions allowing for the shortest tracking time: no extended aspect is excluded beforehand and the robot can possibly visit more than one of them. Initial and final joint positions are not constrained and, therefore, they are part of the solution. The parameters of the algorithm are selected as a result of a trade-off between computation time and accuracy.

The TOTP-R algorithm is executed on the straight line path and the results for different sets of parameters are reported in Table \ref{tab:straight_line_statistics}. PV stands for \emph{pseudo-velocity}, RP for \emph{redundancy parameter}. A necessary condition for global optimality is that at least one actuator saturates for each waypoint. In our DDP algorithm, saturation is never exact because of discretization, thus an actuator is considered in saturation if its velocity, acceleration, jerk, torque or torque rate is above 95\% of its capacity, in agreement with Table \ref{tab:panda_joint_limits}. All solutions of Table \ref{tab:straight_line_statistics} present a bang-bang behavior according to this approximation.

\begin{table}
\centering
\scriptsize
\renewcommand{\arraystretch}{1.2}
\caption{Results of the TOTP-R algorithm on the straight line path}
\label{tab:straight_line_statistics}
\begin{tabular}{ccccccc}
Plan & Number of & PV & RP res. & PV & Cost \\
ID & waypoints & limit & (deg) & res. & (s) \\
\hline
1 & $10$ & $1.4$ & $0.500$ & $0.02$ & $0.655$ \\
2 & $10$ & $1.4$ & $0.250$ & $0.02$ & $0.594$ \\
3 & $10$ & $1.4$ & $0.250$ & $0.01$ & $0.592$ \\
4 & $10$ & $1.4$ & $0.125$ & $0.02$ & $0.574$ \\
5 & $15$ & $1.4$ & $0.125$ & $0.02$ & $0.645$ \\
6 & $20$ & $1.4$ & $0.125$ & $0.02$ & $0.592$ \\
\hline
\end{tabular}
\end{table}

With reference to Table \ref{tab:straight_line_statistics}, we note that the optimal cost consistently is in a neighborhood of $0.6$ s, with a variability of a few milliseconds. By increasing the number of waypoints only (see plans 4-6), the cost may increase or decrease. We should remark that more waypoints correspond to a larger number of constraints in the task and less freedom in deviating from the true straight line. On the other hand, more waypoints allow to reduce the error related to the linear approximation that we make at each step by using a simple Euler integration scheme. Regardless of the number of waypoints selected by the user, we note that, since \eqref{joint_position_sample} is a positional IK scheme, the task space path output by the planner does not differ from the input one. Unlike other planning techniques, task deviations due to, e.g.\ integration of velocity or acceleration equations, are not possible.  As expected, with a fixed number of waypoints, the cost decreases for finer resolutions of the redundancy parameter and the pseudo-velocity (see plans 1-4), although the improvement due to the latter, in our experiments, is negligible (see plans 2 and 3).

\begin{figure}[!ht]
    \centering   
    \includegraphics[width=0.49\textwidth]{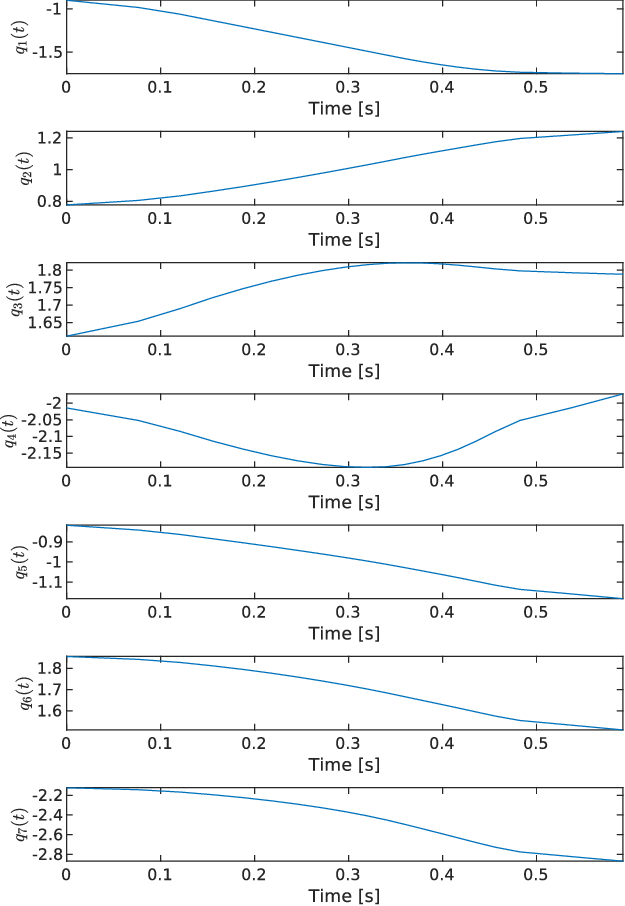}
    \caption{Optimal joint positions for plan 6 of the straight line path.}    
	\label{fig:straight_line_planned_position}
\end{figure}

\begin{figure}[!ht]
    \centering   
    \includegraphics[width=0.49\textwidth]{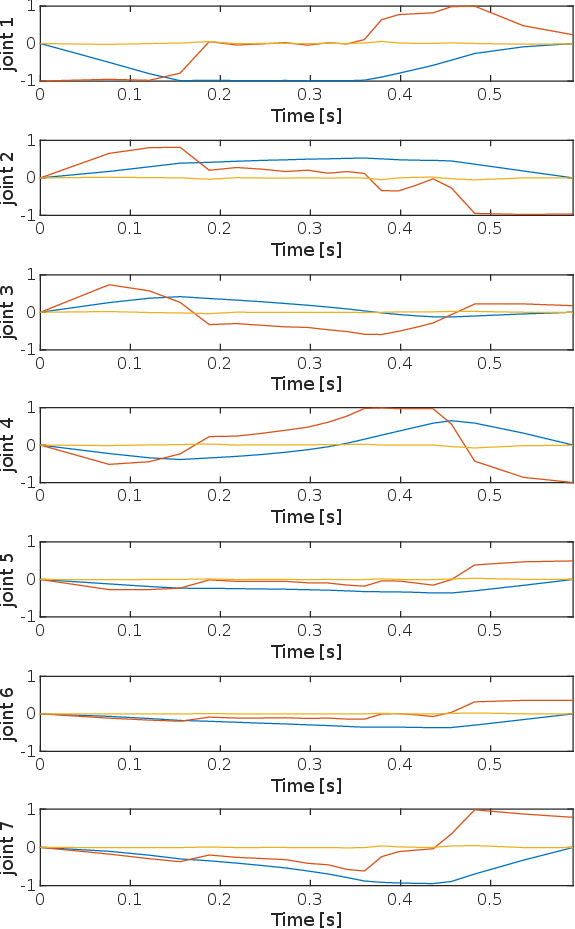}
    \caption{Optimal joint velocities (blue), accelerations (red), and jerks (yellow) for plan 6 of the straight line path, normalized in $[-1,1]$.}    
	\label{fig:straight_line_planned_kinematics}
\end{figure}

\begin{figure}[!ht]
    \centering   
    \includegraphics[width=0.49\textwidth]{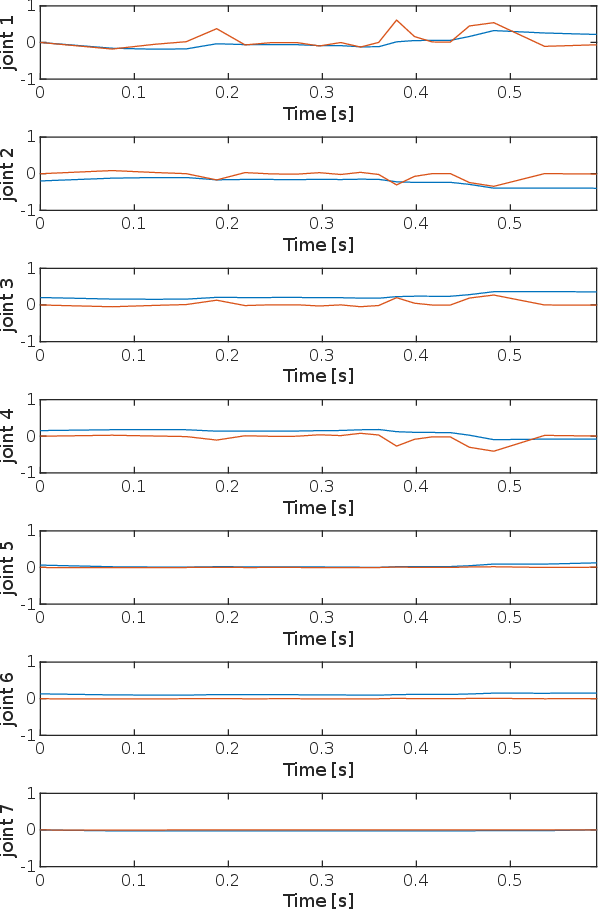}
    \caption{Optimal joint torques (blue) and torque rates (red) for plan 6 of the straight line path, normalized in $[-1,1]$.}    
	\label{fig:straight_line_planned_dynamics}
\end{figure}

The actuation limits of the Panda robot (see Table \ref{tab:panda_joint_limits}), in our experiments, are always such that constraints in \eqref{totpr_velocity_constraint}-\eqref{totpr_acceleration_constraint} activate before constraints in \eqref{totpr_jerk_constraint}-\eqref{totpr_torque_rate_constraint}. This is also due to the fact that planning is performed by considering zero load at the end-effector, while jerk limits are very large. The resolution-optimal joint space solution for plan 6 is reported in Fig.\ \ref{fig:straight_line_planned_position}, its kinematic derivatives in Fig.\ \ref{fig:straight_line_planned_kinematics}, torques and their derivatives in Fig.\ \ref{fig:straight_line_planned_dynamics}. The computation time needed to find such a solution is 219 minutes on a 64-bit Ubuntu 18.04 LTS OS running on an Intel\textsuperscript{\textregistered} Core\textsuperscript{TM} i7-2600 CPU @ 3.40GHz $\times 8$. The TOTP-R algorithm in Fig.\ \ref{alg:totpr_dp_algo} has been implemented in C++ over ROS\footnote{https://www.ros.org/} and MoveIt!\footnote{https://moveit.ros.org/}. No multi-core execution model has been used in the experiments.

We note that, at first, joint 1's acceleration saturates to reach the maximum velocity, which persists for a large portion of the trajectory, between $t=0.16$ s and $t=0.37$ s. Here, there is a rather simultaneous saturation of joint 4's acceleration and joint 7's velocity, up until $t=0.44$ s. In the last segment, joint 1's acceleration saturates again, followed by joint 2's acceleration to perform the final braking. For some segments of the trajectory, two actuators are in saturation at the same time, which is expected for a redundant manipulator, as argued in \cite{Galicki99}.

\begin{figure}[!ht]
    \centering   
    \includegraphics[width=0.5\textwidth]{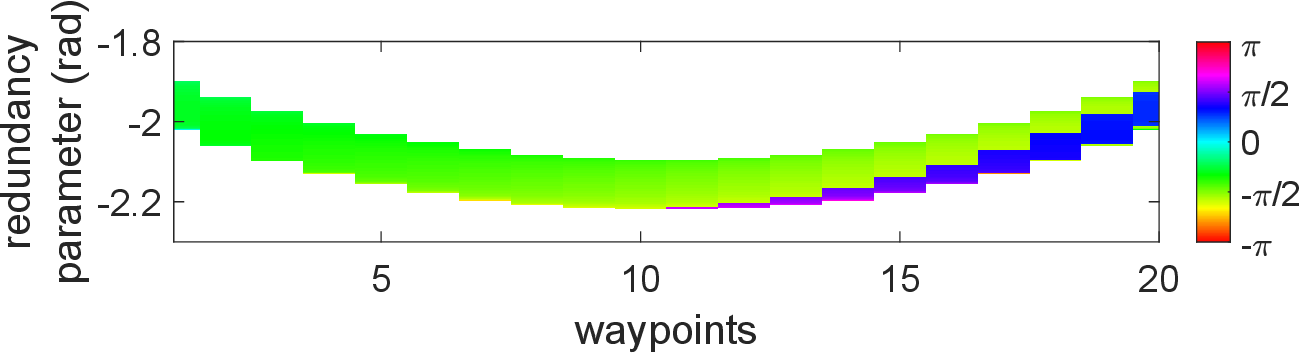}
    \includegraphics[width=0.5\textwidth]{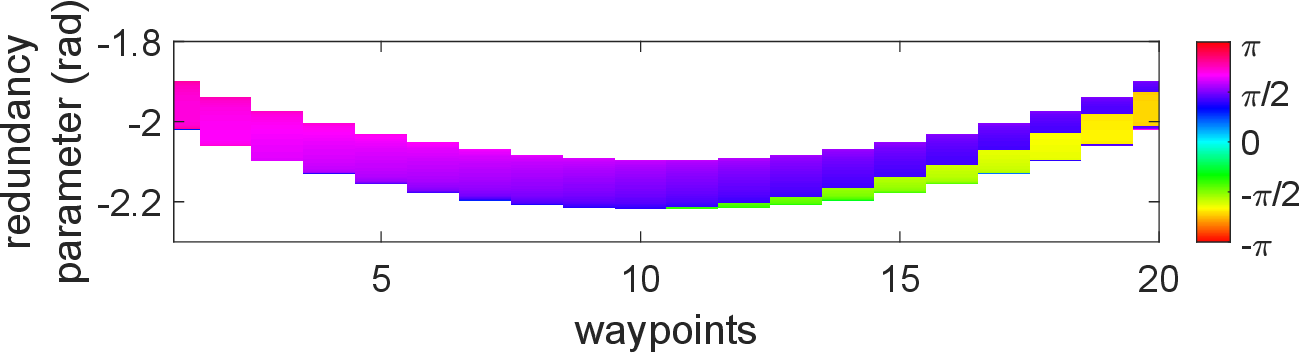}
    \includegraphics[width=0.5\textwidth]{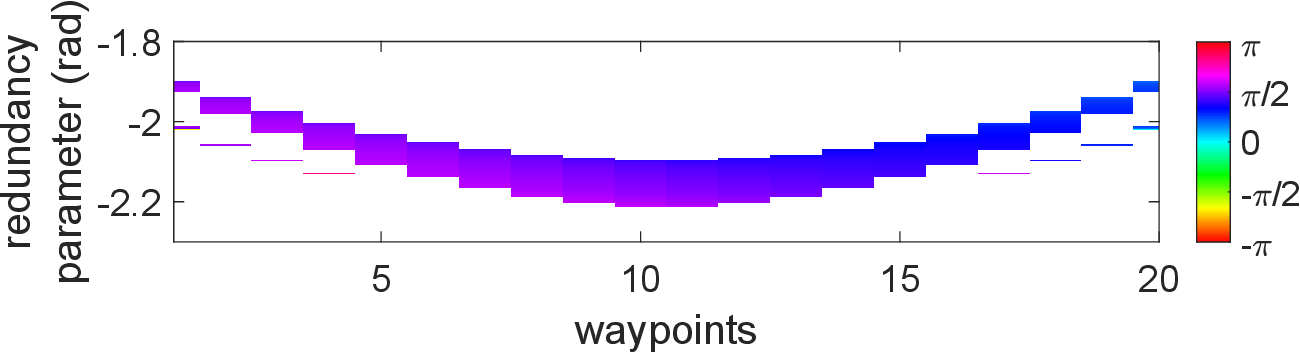}
    \includegraphics[width=0.5\textwidth]{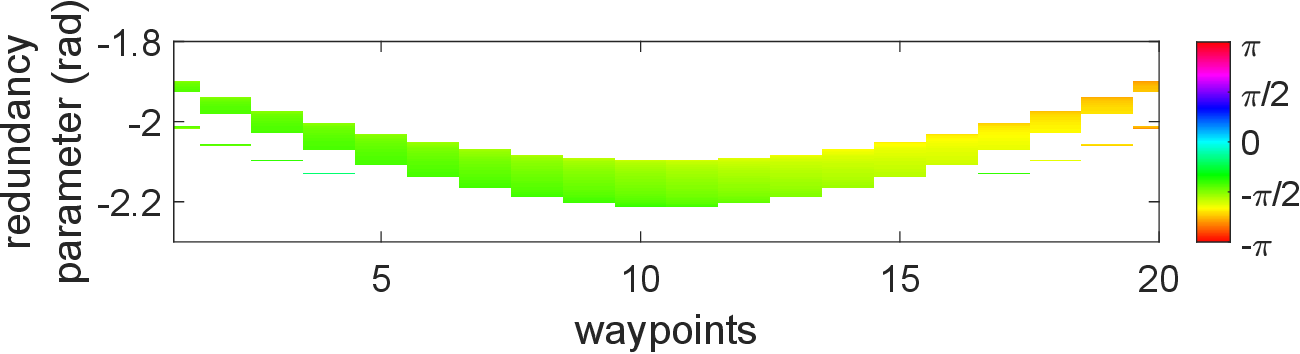}
    \caption{Panda grids for $g=1$ (top) to $g=4$ (bottom), representing $q_1$ for the straight line path; $q_1$ color scale is in radians.}    
	\label{fig:straight_line_ssg}
\end{figure}

A representation of the null-space for the straight line path with 20 waypoints (plan 6) and joint position $q_1$ is reported in Fig.\ \ref{fig:straight_line_ssg} with feasibility maps \cite{FerrentinoJMR, Pamanes}. Because of the task-manipulator geometry and joint limits, only 4 of the 8 grids expected for the Panda have feasible configurations, corresponding to the colored cells. The redundancy parameter, i.e.\ $q_4$, is represented along the y-axis in its physical domain (see Table \ref{tab:panda_joint_limits}). The inverse kinematics solutions are returned by the kinematic solver \emph{IKFast} \cite{Diankov} in a way that grids are not homogeneous.

\begin{figure}[!ht]
    \centering   
    \includegraphics[width=0.49\textwidth]{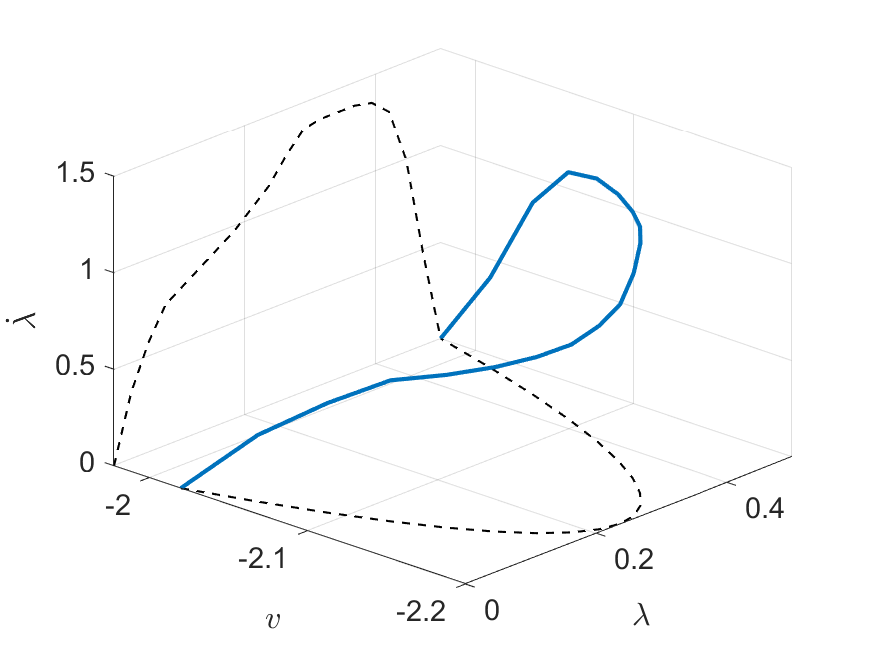}
    \caption{Phase space trajectory (blue) with projections (black) on planes $\lambda$-$\dot{\lambda}$ and $\lambda$-$v$.}    
	\label{fig:straight_line_planned_pst}
\end{figure}

The phase plane trajectory, that in the case of TOTP-R is a \emph{phase space trajectory (PST)}, is reported in Fig.\ \ref{fig:straight_line_planned_pst}. As noted above, for redundant manipulators, the PST is a function of two independent variables that are $\lambda$ (the progress along the path) and $\mathbf{v}$, the redundancy parameter(s). The 3D view provides an indication of how the DDP algorithm exploits the redundancy parameter to increase the pseudo-velocity and consequently compute a better solution with respect to a pre-assigned joint space path.

The best parameters for the straight line path are directly re-used for the ellipse-like path, to assess their consistency across different paths. Since the path is longer ($L = 1.45$ m), a higher number of waypoints should be used to keep a similar spatial resolution. The parameters used for this use case are reported in Table \ref{tab:ellipse_statistics}, together with the optimal cost. Graphs are omitted for the sake of brevity, but similar results are obtained. Even in this case, saturation only happens for velocities and accelerations.

\begin{table}
\centering
\scriptsize
\renewcommand{\arraystretch}{1.3}
\caption{Results of the TOTP-R algorithm on the ellipse-like path with same parameters as plan 6 of Table \ref{tab:straight_line_statistics}.}
\label{tab:ellipse_statistics}
\begin{tabular}{ccccccc}
Length & Number of & PV & RP res. & PV & Cost\\
(m) & waypoints & limit & (deg) & res. & (s)\\
\hline
$1.45$ & $60$ & $1.4$ & $0.125$ & $0.02$ & $1.856$\\
\hline
\end{tabular}
\end{table}

\begin{figure}[!ht]
    \centering   
    \includegraphics[width=0.49\textwidth]{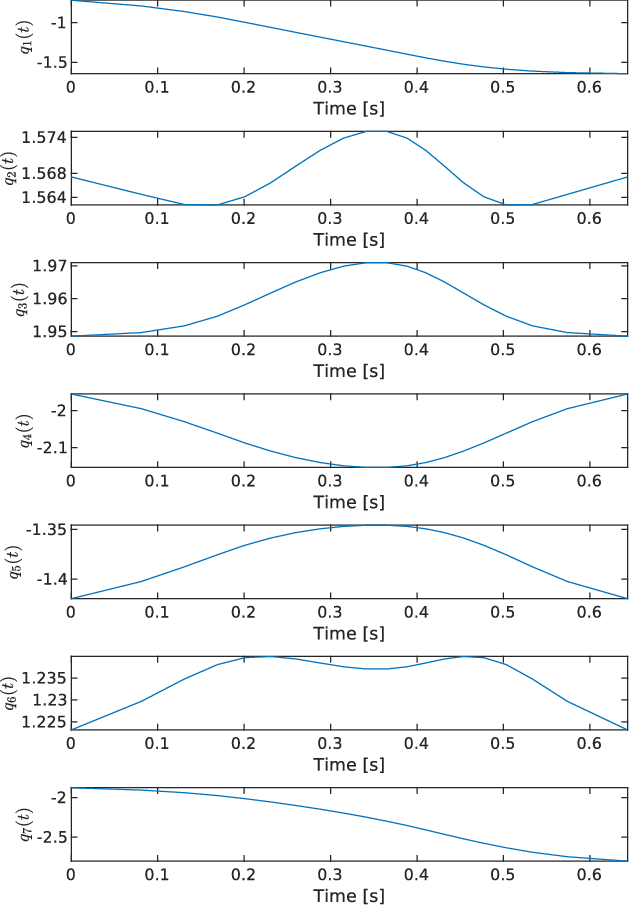}
    \caption{Optimal joint positions for the straight line path, planned with the same parameters as plan 6, with stricter torque limits.}    
	\label{fig:straight_line_restricted_positions}
\end{figure}

\begin{figure}[!ht]
    \centering   
    \includegraphics[width=0.49\textwidth]{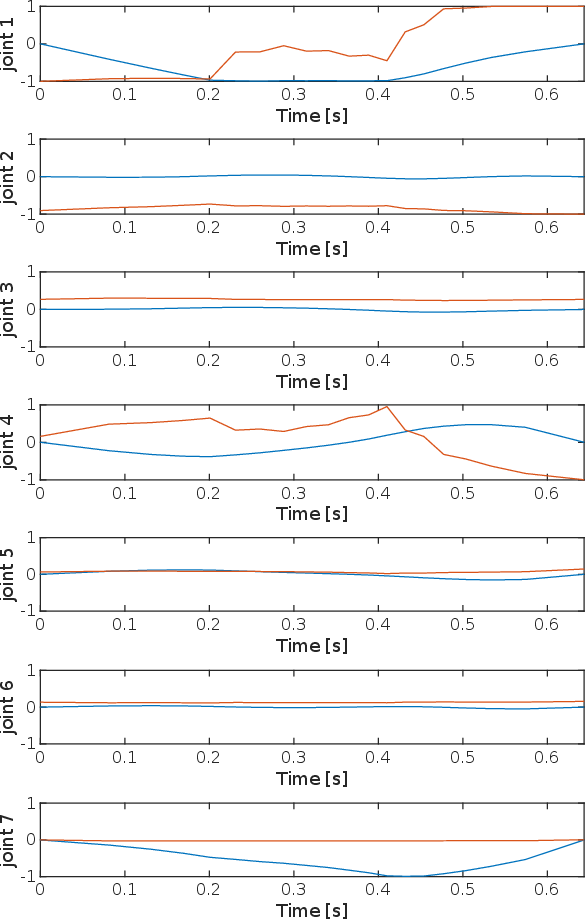}
    \caption{Optimal joint velocities (blue) and torques (red), planned with the same parameters as plan 6, with stricter torque limits, normalized in $[-1,1]$.}
	\label{fig:straight_line_restricted_saturating}
\end{figure}

In order to show that the algorithm is effectively imposing dynamic constraints, e.g.\ actuation torques, we plan the straight line path again by setting stricter torque limits, i.e.\ $\boldsymbol{\tau}_{max} = \left[ 15, 30, 87, 10, 12, 12, 12 \right]^T$, and reusing the configuration parameters of plan 6 in Table \ref{tab:straight_line_statistics}. Because of the stricter limits, the cost of the solution increases to $0.644$ s. Fig.\ \ref{fig:straight_line_restricted_positions} shows the planned joint space trajectory, while Fig.\ \ref{fig:straight_line_restricted_saturating} shows the saturating variables. In particular, joint 1's torque saturates first, until reaching the maximum velocity, which persists until $t = 0.43$ s. Here, joint 7's velocity saturates for a short period of time. In the last segment, we observe a rather simultaneous saturation of joint 1's and joint 2's torques to perform the final braking. Accelerations, jerks and torque rates never saturate.

\subsection{Comparison with a decoupled approach}
\label{sec:comparison_two_stage}

Since we do not have the true minimum time trajectory available, we compare our DP algorithm with a decoupled approach from the literature, in order to assess the gain in performance given by the unified approach proposed in this paper. The same straight line and ellipse-like paths are inverted and time-parametrized with the algorithm in \cite{Chiacchio90}. Each task space waypoint is inverted through the local Jacobian-based iterative optimization scheme
\begin{equation} \label{eq:rr_chiacchio}
{\bf q}_{k+1}(i) = {\bf q}_k(i) + \beta {\bf J}_k^\dagger {\bf e}_k - \alpha ({\bf I}-{\bf J}_k^\dagger {\bf J}_k) \nabla_{\bf q} c_k
\end{equation}
where $k$ is the iteration index, ${\bf J}^\dagger$ is the $n \times m$ full-rank Moore-Penrose pseudo-inverse of the manipulator's geometric Jacobian, ${\bf e}_k$ is the $m \times 1$ pseudo-twist Cartesian error vector, ${\bf I}$ is the identity matrix of appropriate dimension, $\nabla_{\bf q} c_k$ is the gradient of the cost function $c$ selected for redundancy resolution, $\alpha = 10^{-4}$ and $\beta = 10^{-1}$ are gains on the null-space and Cartesian space correction terms, chosen after several trials. The cost is the projection of the dynamic manipulability ellipsoid onto the path's tangent ${\bf t}$, that is associated to the acceleration/deceleration capabilities of the arm along the path \cite{Chiacchio90}:
\begin{equation}
c({\bf q}) = {\bf t}^T({\bf H} {\bf J}^\dagger)^T {\bf H} {\bf J}^\dagger {\bf t}.
\end{equation}
After the joint-space path has been obtained, it is time-parametrized with the algorithm from \cite{Shin86}, by using the same pseudo-velocity limit and a pseudo-velocity resolution of $0.01$. The trajectory tracking times for the straight line and the ellipse-like paths are $1.280$ s and $2.557$ s, about $113\%$ and $38\%$ higher than the unified approach. The CPU time for the whole planning process is about 1-2 minutes. Besides the mere assessment of cost and performance, we remark that the decoupled approach is very sensitive to the parameters $\alpha$ and $\beta$, as well as the specific objective function $c$ selected for redundancy resolution. Other typical choices for $c$ are joint displacement, velocity manipulability ellipsoid projection and multi-objective functions considering two or more performance indices together \cite{Storiale}. Furthermore, the initial joint positions must be assigned and cannot be retrieved as part of the solution. Lastly, the redundancy resolution procedure is not guarantee to converge in case of, e.g., strict joint limits, crossing of singularities. In our solution returning the cost of $1.280$ s (more than twice the average time in Table \ref{tab:straight_line_statistics}), redundancy resolution drives some joints close to their limits, which forces the robot to slow down its progression on the path, reconfigure, and accelerate again. This is the main cause behind the huge deterioration of performance with respect to the unified approach.

\subsection{Comparison with a multiple shooting approach}
\label{sec:comparison_ms}

In Section \ref{sec:related_works}, we identified the technique proposed in \cite{Reiter16a, Reiter16b, Reiter17, Reiter18} as the state of the art in terms of time-optimal planning for redundant robots with a unified approach. We discussed some of its advantages, including the possibility of easily adding generic, possibly nonlinear constraints on the joint positions and their derivatives and generating smooth solutions. Here, we compare it with our approach on the straight line path. The problem formulation is:
\begin{equation} \label{eq:ms}
\begin{split}
\min_{\bf y}~& t_f \\
\mbox{s.t.}~& 0 \leq \lambda_k \leq 1 \\
& \dot{\lambda}_k \geq 0 \\
& {\bf q}_{min} \leq {\bf q}_k \leq {\bf q}_{max} \\
& \dot{\bf q}_{min} \leq \dot{\bf q}_k \leq \dot{\bf q}_{max} \\
& \ddot{\bf q}_{min} \leq \ddot{\bf q}_k \leq \ddot{\bf q}_{max} \\
& \boldsymbol{\tau}_{min} \leq \boldsymbol{\tau}_k \leq \boldsymbol{\tau}_{max} \\
& \ddot{\bf q}_k = {\bf J}_k^\dagger(\dot{\bf w}_k - \dot{\bf J}_k \dot{\bf q}_k) + \sum_{i=1}^{r} \gamma_{k,i} {\bf a}_{k,i} \\
& \boldsymbol{\tau}_k = {\bf H}({\bf q}_k) \ddot{\bf q}_k + {\bf f}({\bf q}_k, \dot{\bf q}_k) \\
& {\bf y}_{k+1} - {\bf g}_{RK4}({\bf y}_k, {\bf u}_k) = {\bf 0} \\
& {\bf x}_k = {\bf k}({\bf q}_k) \\
& \lambda_0 = 0, \lambda_N = 1 \\
& \dot{\lambda}_0 = \dot{\lambda}_N = 0 \\
& \dot{\bf q}_0 = \dot{\bf q}_N = {\bf 0}
\end{split}
\end{equation}
where $N$ is the number of multiple shooting intervals, $k=1,\ldots,N$, ${\bf y} = \left[ {\bf y}^T_0, {\bf u}^T_0, \ldots, {\bf y}^T_{N-1}, {\bf u}^T_{N-1}, {\bf y}^T_{N}, t_f\right]^T$ is the decision vector, with ${\bf y}_k = \left[ \lambda_k, \dot{\lambda}_k, {\bf q}^T_k, \dot{\bf q}^T_k \right]^T$, i.e.\ the state vector at the $k$-th interval, ${\bf u}_k = \left[ \ddot{\lambda}_k, \gamma_{k,1}, \ldots, \gamma_{k,r}\right]^T$, i.e.\ the input vector at the $k$-th interval,  ${\bf w}_k \in \mathbb{R}^m$ is the end-effector twist in the base frame, ${\bf a}_{k,i}$ are the $r$ vectors in the null space basis of ${\bf J}_k$, $\gamma_{k,i}$ are their gains, and ${\bf g}_{RK4}$ is the 4-th order explicit Runge-Kutta integration scheme, employed to impose continuity between multiple shooting intervals.

We select $N=20$, yielding 359 decision variables in ${\bf y}$, initialize ${\bf y}_k$ and $\ddot{\lambda}_k, \forall k$, to match the joint space trajectory obtained from Section \ref{sec:comparison_two_stage}, and we set $\gamma_{k} = \gamma_{k,1} = 0, \forall k$, as done in \cite{Reiter16a, Reiter16b}. Since the algorithms in \cite{Reiter16a, Reiter16b, Reiter17, Reiter18} are not publicly available, we implement problem \eqref{eq:ms} in MATLAB and solve it with \texttt{fmincon} using the interior-point method. The trajectory tracking time is $1.165$ s (about $94\%$ higher than our approach), the CPU time is 192 minutes. The involved quantities only partially saturate, which confirms the locally-optimal nature of the solution. Indeed, as noted in \cite{Reiter16a, Reiter16b, Reiter17, Reiter18}, global optimality is not guaranteed as problem \eqref{eq:ms} is non-convex, and neither is convergence for some choices of the parameters and initial guess.

\subsection{Additional comparison trials}
\label{sec:additional_trials}

In order to confirm the correctness of the comparison, we ran additional trials assuming availability of a-priori information on the optimal solution of the straight line use case. In particular, by initializing problem \eqref{eq:rr_chiacchio} with the initial positions retrieved by our DDP algorithm (Fig.\ \ref{fig:straight_line_planned_position}), setting $c$ to joint displacement \cite{Storiale}, and re-executing the steps of Section \ref{sec:comparison_two_stage} again, a trajectory tracking time of $0.628$ s is achieved. If we use the latter solution as the initial guess for ${\bf y}$ in \eqref{eq:ms}, the interior-point algorithm is able to consistently reach trajectory tracking times of $\sim0.600$ s, for different configurations of the parameters $N, \gamma_k$. This confirms the strong dependency of the approaches in Sections \ref{sec:comparison_two_stage} and \ref{sec:comparison_ms} on the initial guess, as well as their locally-optimal nature. In our approach, no initial information on the solution needs to be available, as a global exploration of the search space is performed.

\subsection{Execution}
\label{sec:execution}

In this section, we assess the feasibility of the generated plan. We want to confirm that, despite the discrete approximations embedded in the DDP algorithm, the plan is actually executable through a standard interpolation and control scheme. We experimentally show that no additional requirements are imposed to the control system, besides controlling the plant in proximity of actuation limits.

The execution of the TOTP-R algorithm on either trajectory generates several control signals, as seen in Fig.\ \ref{fig:straight_line_planned_kinematics} and Fig.\ \ref{fig:straight_line_planned_dynamics}. Before they can be used as commands for the Panda robot, it is necessary to interpolate them at the frequency of the controller, i.e.\ 1 kHz. We choose to control joint positions, while command torques are generated with an internal joint impedance controller, configured with maximum stiffness. Position control has already been used in other works, e.g.\ \cite{AlKhudir}, dealing as well with execution of time-optimal joint space solutions, with satisfactory results. Furthermore, it is a typical choice as many robots do not allow for torque commands for practicality and safety reasons.

\begin{figure}[!ht]
    \centering   
    \includegraphics[width=0.49\textwidth]{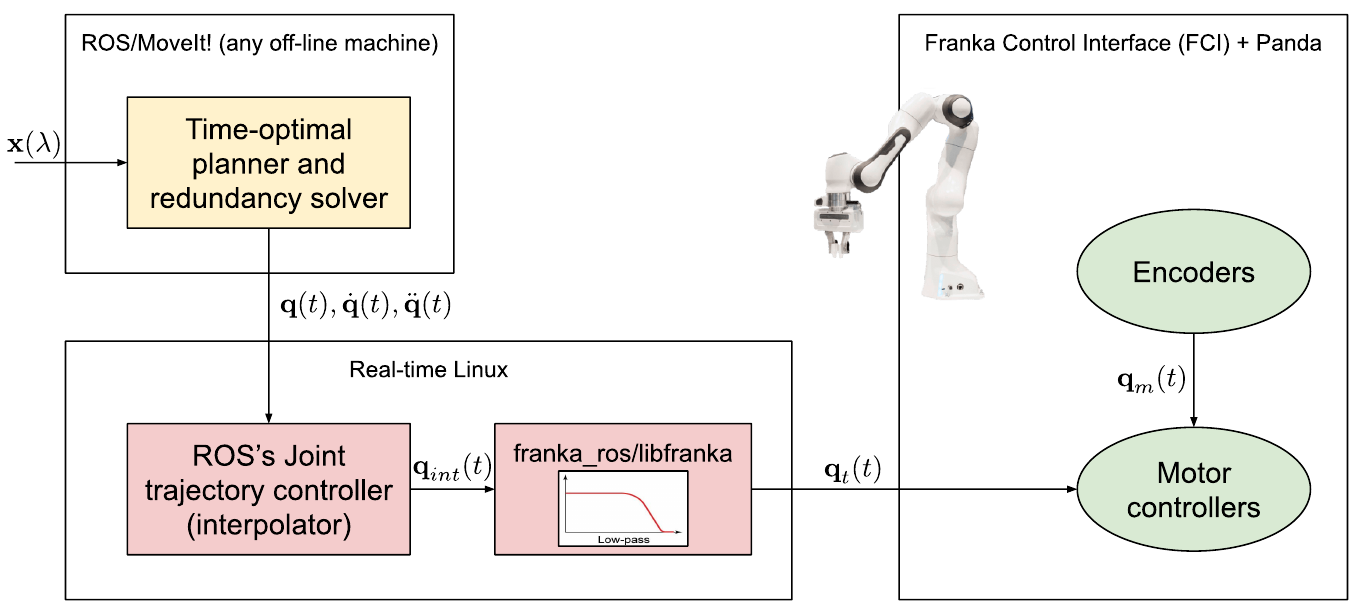}
    \caption{Block diagram showing the control setup.}    
	\label{fig:panda_control_setup}
\end{figure}

A high-level description of the control setup is shown in Fig.\ \ref{fig:panda_control_setup}. As mentioned, Algorithm \ref{alg:totpr_dp_algo} is implemented in ROS/MoveIt!\ and deployed as a node. Task space paths ${\bf x}(\lambda)$ are provided in bag files and optimal joint space solutions, i.e.\ curves ${\bf q}(t)$, $\dot{\bf q}(t)$ and $\ddot{\bf q}(t)$ are also stored in bag files. The controller manager of \texttt{ros\_control}\footnote{http://wiki.ros.org/ros\_control} is configured to work with the Panda robot hardware interface (provided through \texttt{franka\_ros}\footnote{https://frankaemika.github.io/docs/franka\_ros.html}) and to load a robot-agnostic joint trajectory controller (JTC). Internally, the JTC performs quintic interpolation on the input signals to deliver smoother signals to the robot driver.

In order to guarantee that the interpolation of the planned trajectories still generates curves that are within bounds, we exploit some basic signal processing functions that are already provided by the manufacturer, i.e.\ a low-pass filter and a rate limiter. In particular, both of them are active to deliver less oscillating commands and ensure compliance to actuator limits. The employment of a filter to smooth the trajectory is also beneficial to eliminate the high frequency content that excites the unmodeled joint elasticity that characterizes joint transmissions in time-optimal control \cite{Kim19}. In fact, we note that controlling the motors in the proximity of their limits, so that they saturate, reserves no margin for the controller to compensate for unmodeled dynamics and/or uncertainties in the modeled dynamics. Some techniques explictly addressing this control issues can be found in \cite{Arai, Dahl90, Shiller95, Shiller96, Kieffer, Dahl94, Moreno-Valenzuela, Eom, Niu, Tam}.

\begin{figure}[!ht]
    \centering   
    \includegraphics[width=0.49\textwidth]{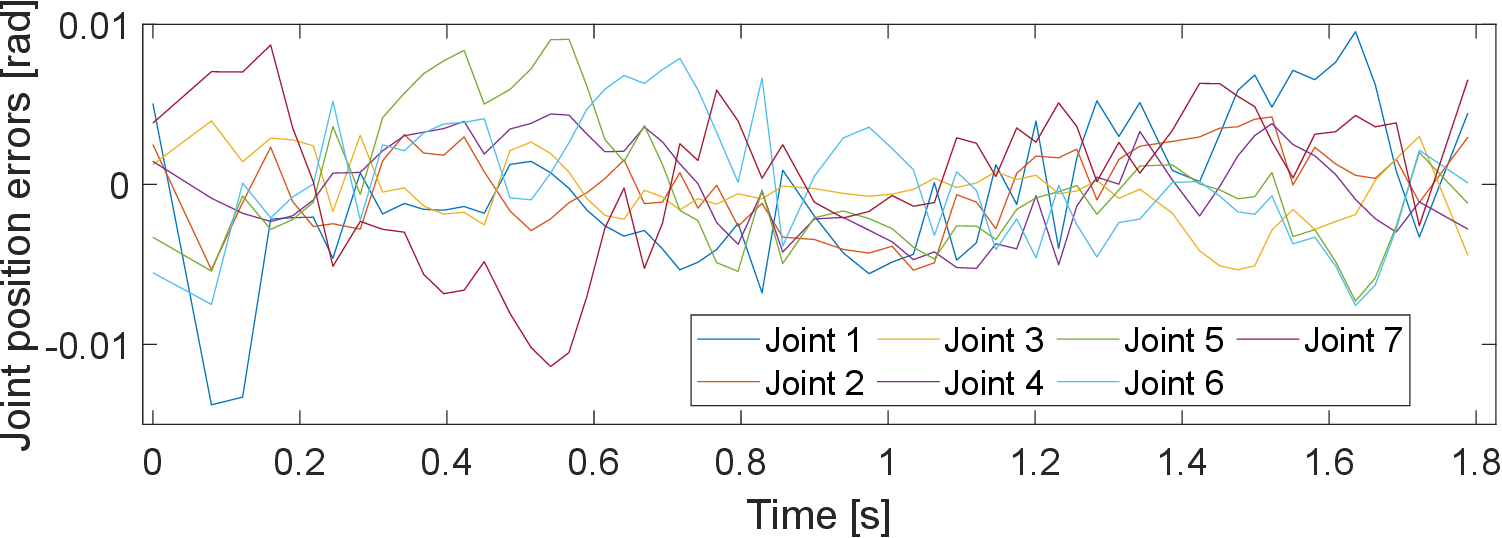}
    \caption{Joint position errors for the ellipse-like trajectory.}
	\label{fig:ellipse_planned_vs_actual}
\end{figure}

The joint position tracking errors for the ellipse-like trajectory are reported in Fig.\ \ref{fig:ellipse_planned_vs_actual}. The robot completes the motion in $1.859$ s, $0.13$\% slower than planned with a maximum error in joint space of $0.8$ deg. The delay introduced by the low-pass filter has been removed in the plots to provide a faithful comparison with the planned references. A video of the execution\footnote{\url{https://t.ly/ZKelW
}} is provided as supplemental material. While the actual tracking performance is dependant on the specific robot controller, we show that the trajectories planned with our methodology are feasible on a general-purpose architecture, i.e.\ the controller does not abort because of, e.g., out-of-bound or jerky references, poor smoothness or excessive trajectory tracking error. Moreover, the latter is comparable to other experiments of time-optimal trajectories execution \cite{AlKhudir}. In comparing with previous works, we should remark that our plans are calculated with the real robot capacities, as reported in Table \ref{tab:panda_joint_limits}, while conservative values might be employed elsewhere, e.g.\ a given percentage of the actual limit.

\begin{figure}[!ht]
    \centering   
    \includegraphics[width=0.49\textwidth]{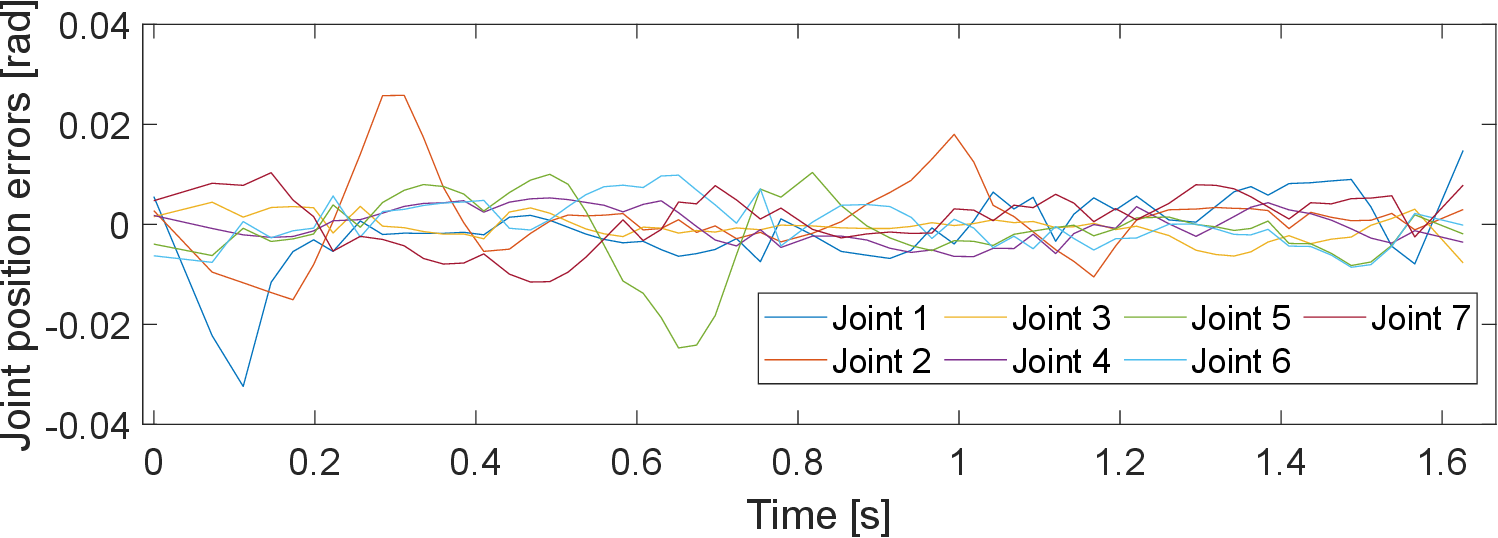}
    \caption{Joint position errors for a trajectory $10$\% faster than the optimum.}
	\label{fig:ellipse_planned_vs_actual_faster}
\end{figure}

The same planned trajectory is scaled in time to be $10$\% faster and is sent to the robot again to estimate its performance. In this case, the motion is completed in $1.70$ s, $0.67$\% slower than planned, with a maximum error of $1.86$ deg, associated to joint 1 at $t=0.11$ s. Joint position tracking errors are reported in Fig.\ \ref{fig:ellipse_planned_vs_actual_faster}. As expected, the robot cannot be controlled beyond its capacities and the actual motion results in a remarkable degradation of tracking performances. We also note here that the controller aborts for trajectories that are more than $10$\% faster than the optimal plan.

\section{Conclusion}
\label{sec:conclusion}

In this paper, a DDP framework has been presented to plan trajectories for redundant robots along prescribed paths. The framework unifies redundancy resolution and trajectory planning so as to solve them together to effectively utilize kinematic redundancy to optimize the objective function defined at trajectory planning level. The extreme flexibility of DDP has been exploited to accommodate generic constraints that characterize real applications. DP ensures resolution-optimality, i.e.\ the asymptotic achievement of the globally-optimal solution as the dicretization step tends to zero.

Although it is perfectly suited to solve one or the other problem separately, and it is applicable to many offline planning scenarios, e.g.\ planning of cyclic operations, DDP still requires hours to days to solve a medium-complexity planning problem for redundant robots (one or two orders of magnitude higher than other approaches) and, most importantly, its scalability for systems characterized by a higher degree of redundancy, i.e.\ $r>1$, is questionable. Also, in order to impose exact constraints on higher-order derivatives, the phase space should be further augmented with additional dimensions. 

Nevertheless, the proposed state space formalization can be employed with optimization techniques other than DDP, such as randomized algorithms, for those scenario where planning time is a constraint. Additional future developments might regard a deeper study into decoupled techniques to evaluate whether some specific objective functions can be used for redundancy resolution (e.g., acceleration capability of the manipulator) that allow optimizing trajectory planning at a later stage. Also, it should be investigated whether the optimization of integral performance indices can be beneficial to this respect, in place of the local optimization methods discussed in \cite{Chiacchio90, Basile, AlKhudir}.

\section*{Acknowledgments}
The authors would like to thank Simon Lacroix for kindly supporting the setup of the experiments, and Federica Storiale and Giuseppe Santoriello for contributing to some of the software modules employed in the experiments.


\bibliographystyle{IEEEtran}
\bibliography{IEEEabrv,tase-time-optimal-redundant}
 
\vspace{11pt}

\begin{IEEEbiography}[{\includegraphics[width=1in,height=1.25in,clip,keepaspectratio]{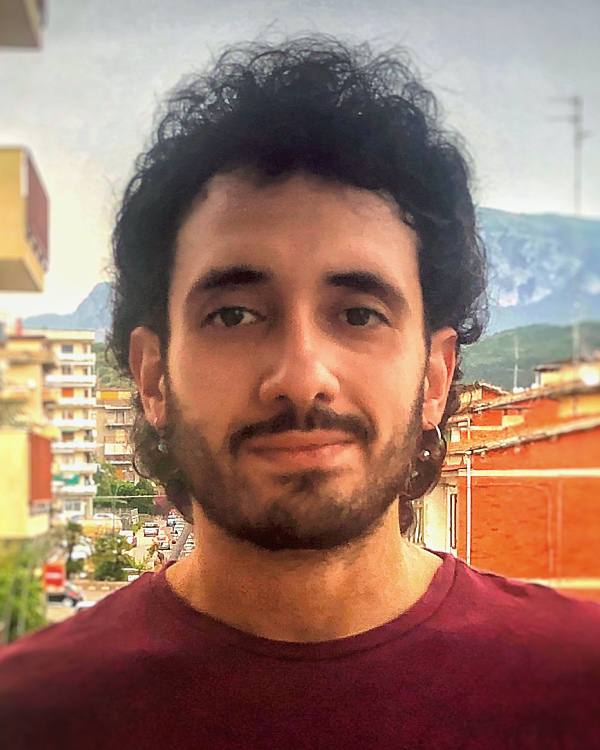}}]{Enrico Ferrentino} (M'21-S'20)
received the M.S.\ degree in computer and automation engineering from the Polytechnic University of Turin, Turin, Italy and the Ph.D.\ degree in robotics from the University of Salerno, Fisciano, Italy. Between 2013 and 2014, he was Visiting Scholar at the NASA Jet Propulsion Laboratory in Pasadena, USA, where he joined the Axel tethered robot project, in the Robotic Mobility group led by Dr.\ I.\ Nesnas
Between 2014 and 2017 he was Ground Segment Engineer at ALTEC S.p.A., Turin, Italy, where he joined the ESA robotic mission ExoMars.
In 2019, he was Visiting Scholar at LAAS-CNRS in Toulouse, France, where he joined the H2020 PRO-ACT project, in the team led by Prof. A. Franchi.
Since 2020, he is Lecturer of Robotics and Medical Robotics with the Department of Information and Electric Engineering and Applied Mathematics at the University of Salerno and, since 2021, Researcher of the Automatic Control Group in the same institution. His research interests include optimal planning of redundant and cooperating robots, interaction control, human-robot collaboration. His applications of interest fall in the fields of industrial, aerospace and medical robotics.
\end{IEEEbiography}

\vspace{11pt}

\begin{IEEEbiography}[{\includegraphics[width=1in,height=1.25in,clip,keepaspectratio]{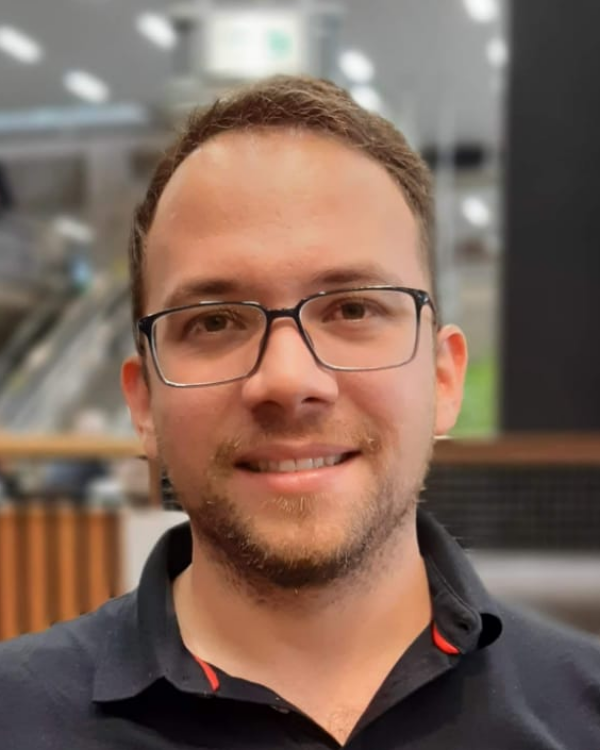}}]{Heitor J. Savino}
has a B.S.\ degree in Mechatronics Engineering from Amazon State University (UEA), a M.Eng.\ from Amazon Federal University (UFAM), and a Doctoral degree in Electrical Engineering from Federal University of Minas Gerais (UFMG), 2016. 

Visiting graduate student at Massachusetts Institute of Technology (MIT), as member of Interactive Robotics Group (IRG), 2015-2016. Conducted post-doctoral research at UFMG from 2016-2017, and at LAAS-CNRS, 2019-2020, as part of the H2020 PRO-ACT project. Assistant Professor at Federal University of Alagoas (UFAL) 2017-2021, at Institute of Computing (IC). 
Since 2022, is the founder and leading researcher at Ambev Robotics Laboratory, subsidiary of Anheuser-Busch InBev, to conduct research on robotics for the beverage industry.
\end{IEEEbiography}

\vspace{11pt}

\begin{IEEEbiography}[{\includegraphics[width=1in,height=1.25in,clip,keepaspectratio]{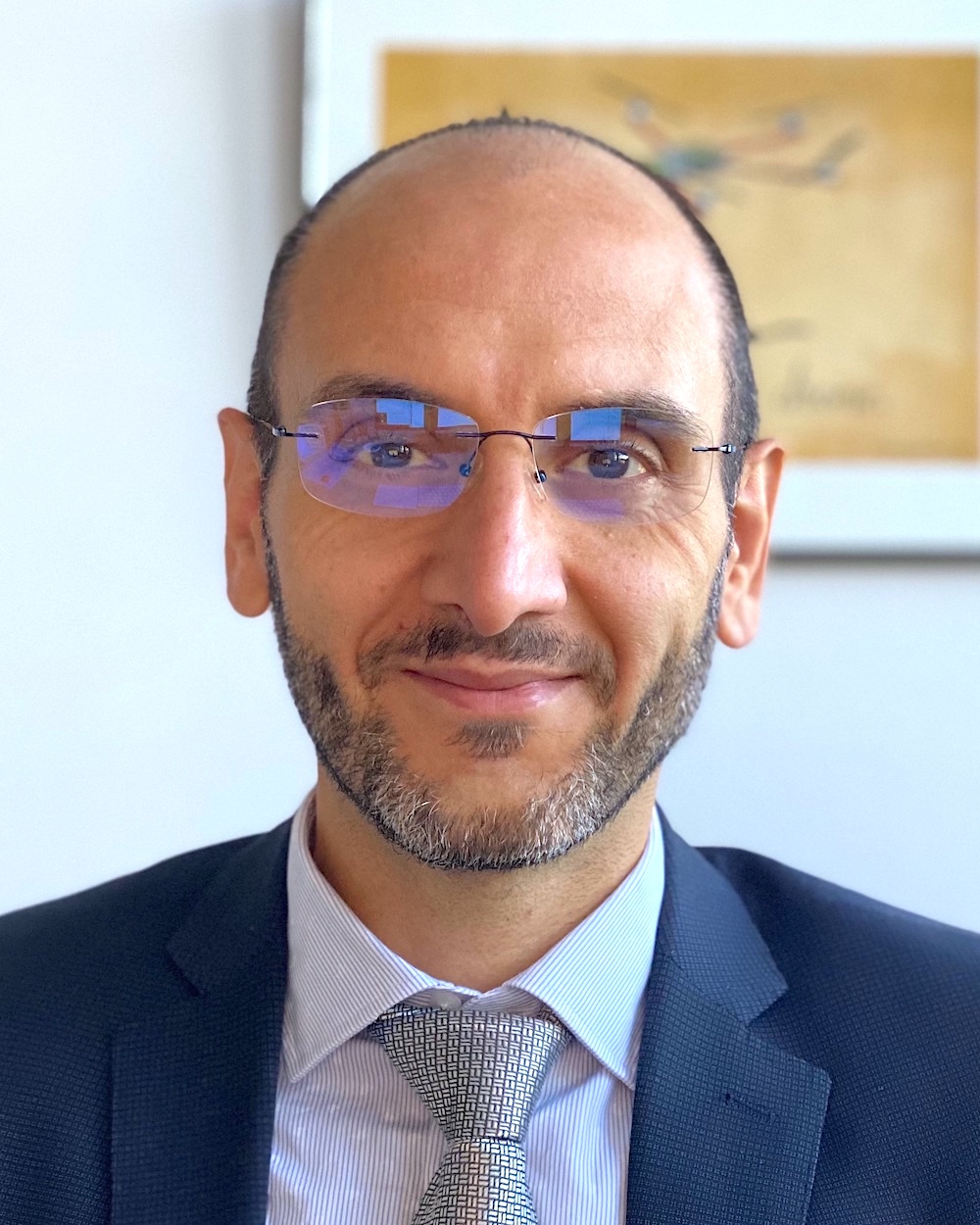}}]{Antonio Franchi} (F'23--SM'16--M'07) received the M.Sc.~degree in electrical engineering and the Ph.D.~degree in system engineering from the Sapienza University of Rome, Rome, Italy, in 2005 and 2010, respectively, and the HDR~degree in Science, from the  National Polytechnic Institute of Toulouse in 2016. From 2010 to 2013 he was a research scientist at the Max Planck Institute for Biological Cybernetics, T\"ubingen, Germany. From 2014 to 2019 he was a tenured CNRS researcher at LAAS-CNRS, Toulouse, France. 
From 2019 to 2021 he was an associate professor at the University of Twente. Enschede, The Netherlands.
Since 2022 he is a full professor in aerial robotics control at the University of Twente. Enschede, The Netherlands. Since 2023 he is also a full professor 
at the department of Computer, Control and Management Engineering, Sapienza University of Rome, Rome, Italy.
He co-authored more than 160 peer-reviewed international publications on design and control of robotic systems applied to multi-robot systems and aerial robots.
\end{IEEEbiography}

\vspace{11pt}

\begin{IEEEbiography}[{\includegraphics[width=1in,height=1.25in,clip,keepaspectratio]{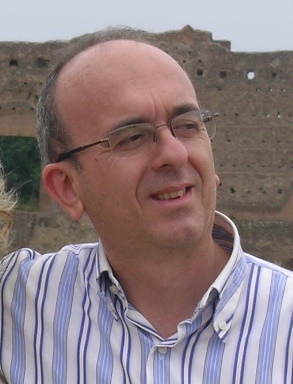}}]{Pasquale Chiacchio} is Professor of Automatic Control and Robotics in the Department of Information and Electric Engineering and Applied Mathematics, University of Salerno. His main research interests include robotics and modeling, and control of discrete event systems. In the robotics field, he has been working on robot control and identification, inverse kinematic problem, interaction control, control of redundant manipulators, control of cooperative manipulators. In the discrete event systems field, he has been working on supervisory control based on monitors, optimal supervisory control, and formal specification for supervisory systems. The results have been published in the main journals of the sector and have been accompanied by an intense experimental activity. He has been the coordinator of two Research Projects of National Interest (PRIN) and has been involved in research projects funded by the European Union (ECHORD, AIRobots, EuRoC, LOCOMACHS, LABOR). He is Senior member of the Institute of Electrical and Electronic Engineers (IEEE) and member of the Italian Society of Control Researchers. In December 2011, he was nominated Knight, then in July 2014 promoted to Officer of the Order of Merit of the Italian Republic. From 2016 he is the Director of the Ph.D. program in Information Engineering.
\end{IEEEbiography}

\vfill

\end{document}